%% file: ms.tex
\def\year{2021}\relax
\documentclass[letterpaper]{article} 
\usepackage{aaai21_arxiv}  
\usepackage{times}  
\usepackage{helvet} 
\usepackage{courier}  
\usepackage[hyphens]{url}  
\usepackage{graphicx} 
\usepackage{natbib}  
\usepackage{caption} 
\nocopyright
\usepackage{comment}
\usepackage{subcaption}
\usepackage{amsmath,amssymb} 
\usepackage{color}

\usepackage[export]{adjustbox}
\usepackage{graphics}
\usepackage{booktabs}
\usepackage{makecell}
\usepackage{multirow}
\usepackage{xcolor,soul,colortbl}
\usepackage{sidecap}
\usepackage{footnote}
\makesavenoteenv{tabular}
\makesavenoteenv{table}

\usepackage{enumitem}
\usepackage{bbold}

\newcommand{\PP}[1]{\textcolor{black}{#1}}

\newcommand{\DHh}[1]{\textcolor{black}{#1}}
\newcommand{\AV}[1]{\textcolor{black}{#1}}

\newcommand\blfootnote[1]{%
  \begingroup
  \renewcommand\thefootnote{}\footnote{#1}%
  \addtocounter{footnote}{-1}%
  \endgroup
}

\title{Artificial Dummies for Urban Dataset Augmentation}
\author{
    Anton\'in Vobeck\'y\textsuperscript{\rm †}, David Hurych\textsuperscript{\rm ‡}, Michal U\v ri\v c\' a\v r\textsuperscript{\rm §}, Patrick P\'erez\textsuperscript{\rm ‡}, Josef \v Sivic\textsuperscript{\rm †} \\
}
\affiliations {
    \textsuperscript{\rm †}CIIRC CTU, 
    \textsuperscript{\rm ‡}valeo.ai, 
    \textsuperscript{\rm §}Independent researcher
}

\begin{document}

\maketitle

\begin{abstract}
\input{include/abstract}
\end{abstract}

\input{include/introduction}
\input{include/related}
\input{include/method}
\input{include/experiments}
\input{include/experiment_nightowls}
\input{include/experiment_csp}
\input{include/nightowls_detection}
\input{include/conclusion}

\input{include/acknowledgements}

\clearpage

\bibliography{ms}

\clearpage

\appendix
\title{Artificial Dummies for Urban Dataset Augmentation: Appendix}
\maketitlesupp
\renewcommand{\thesection}{\Alph{section}}
\input{supplementary_raw}

\end{document}

%% file: include/abstract.tex
Existing datasets for training pedestrian detectors in images suffer from limited appearance and pose variation. The most challenging scenarios are rarely included because they are too difficult to capture due to safety reasons, or they are very unlikely to happen. The strict safety requirements in assisted and autonomous driving applications call for an extra high detection accuracy also in these rare situations. Having the ability to generate people images in arbitrary poses, with arbitrary appearances and embedded in different background scenes with varying illumination and weather conditions, is a crucial component for the development and testing of such applications.
The contributions of this paper are three-fold.
First, we describe an augmentation method for controlled synthesis of urban scenes containing people, thus producing rare or never-seen situations. This is achieved with a data generator (called DummyNet) with disentangled control of the pose, the appearance, and the target background scene.
Second, the proposed generator relies on novel network architecture and 
associated loss that takes into account the segmentation of the foreground person and its composition into the background scene.
Finally, we demonstrate that the data generated by our DummyNet improve performance of several existing person detectors across various datasets as well as in challenging situations, such as night-time conditions, where only a limited amount of training data is available. 
In the setup with only day-time data available, we improve the night-time detector by $17\%$ log-average miss rate over the detector trained with the day-time data only.
\footnote{Code is available at \url{https://github.com/vobecant/DummyNet}}
\blfootnote{† Czech Institute of Informatics, Robotics and Cybernetics at the Czech Technical University in Prague.}

%% file: include/introduction.tex
\section{Introduction} \label{sec:introduction}

\begin{figure*}[h!]
\centering
\includegraphics[width=0.99\linewidth]{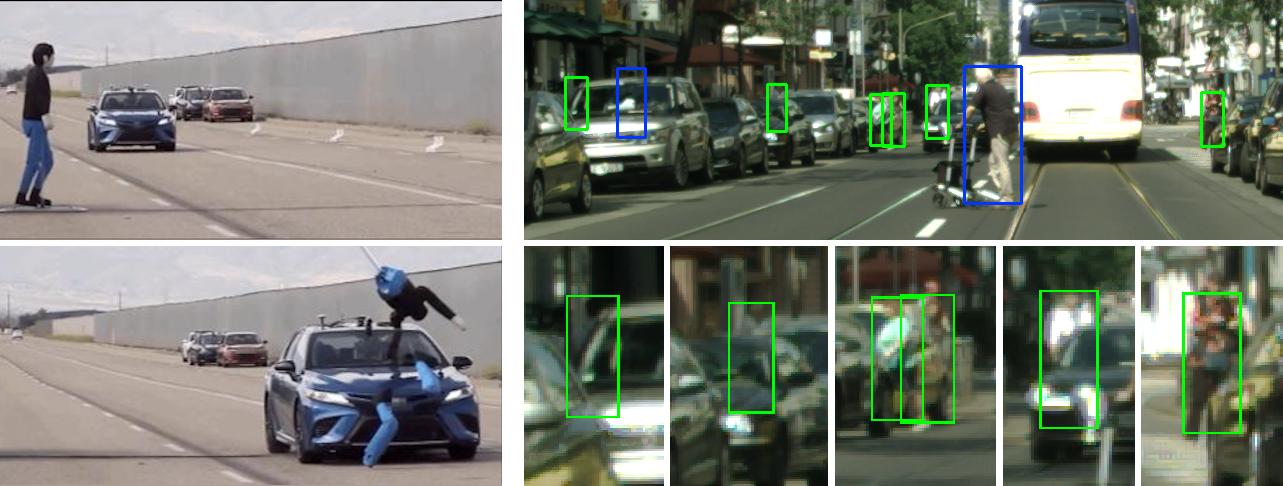}
\vspace{-1.5ex}
\caption{\small {\bf Left: Testing emergency breaking with a standard pedestrian dummy} with a fixed pose and unified clothing. {\bf Right:} Illustration of the {\bf improved performance of state-of-the-art person detector} (``Center-and-Scale-Prediction'', CSP,~\cite{liu2019high}) trained on a dataset set augmented with samples produced by our DummyNet generator. Top: full testing scene. Bottom: close-ups. Detections (all correct) produced by both vanilla and DummyNet augmented CSP person detector are in blue. Additional detections (all correct) produced only by the DummyNet augmented CSP detector are shown in green. Note that the additional detections often correspond to hard occluded examples and would be missed by the state-of-the-art CSP person detector.
}
\label{fig:teaser}
\vspace{-3.5ex}
\end{figure*}

A high-quality dataset is a crucial element for every system using statistical machine learning and should be a representative sample of the target scenarios. Usually, however, we do not have access to such high-quality data due to limited resources for capture and annotation or the inability to identify all the aspects of the target use-case in advance. A key challenge is to cover all the corner-case scenarios that might arise to train and deploy a reliable model.

Dataset diversity is especially important for the automotive industry where handling highly complex situations in a wide variety of conditions (weather, time of day, visibility, etc.) is necessary given the strict safety requirements.
Our goal is to address these requirements and enable a controlled augmentation of datasets for pedestrian detection in urban settings for automotive applications. 
While we focus here only on \textit{training} dataset augmentation as a first, crucial step, the approach also aims at generating data for the system \textit{validation}. Validation on such augmented data could complement track tests with dummy dolls of unrealistic appearance, pose, and motion, which are the current industry standard
(Fig.\,\ref{fig:teaser}). By analogy, we named our system DummyNet after this iconic doll that replaces real humans.

The key challenge in automotive dataset augmentation is to enable sufficient control over the generated distributions via input parameters that describe important corner cases and missing situations. In this work, we take a step in that direction and develop a method for controlled augmentation of person datasets, where people with adjustable pose and appearance are synthesized into real urban backgrounds in various conditions. This is achieved by a new Generative Adversarial Network (GAN) architecture, coined DummyNet, which takes as input the desired person pose, specified by skeleton keypoints, the desired appearance, specified by an input image,  and a target background scene. See the image synthesis diagram in Fig.\,\ref{fig:training_scheme}. The output of DummyNet is the given background scene containing the pedestrian with target pose and appearance composited into one image.

We demonstrate that augmenting training data in this way improves person detection performance, especially in low-data regimes where the number of real training examples is small or when the training and target testing distributions differ (e.g., day/night). The basic assumption of matching training and testing distributions is typically not satisfied when developing real detection systems (e.g., different country, time of day, weather, etc.). Our method allows for adapting the model to known or estimated target (real world) distribution via controlled augmentation of the training data.

\vspace{-1ex}
\paragraph{Contributions.}
Our contributions are three-fold: (1) we develop an approach (DummyNet) for controlled data augmentation for person detection in automotive scenarios that enables independent control of the person's \emph{pose}, \emph{appearance} and the \emph{background} scene; 
(2) the approach relies on a novel architecture and associated \emph{appearance loss} that take into account the segmentation of the foreground pedestrian and its composition into the background scene.
(3) we demonstrate that the data generated by our DummyNet improve several existing person detectors \AV{with different architectures} on standard benchmarks, including both day-time as well as challenging night-time conditions.
In the setup with only day-time data available, using our artificial data, we improve the night-time detector by $17\%$ LAMR over the detector trained with the day-time data only.

%% file: include/related.tex
\section{Related work \PP{and positioning}} \label{sec:related}

\paragraph{Inserting humans.}
Inserting humans into a given scene is an old problem in image editing and visual effects. There are multiple good tools for image/video clipart~\cite{lalonde-siggraph-07} and for producing high-quality images~\cite{Karsch:2011:RSO:2024156.2024191,Kholgade:2014:OMS:2601097.2601209}. However, such methods are meant for interactive manipulation of individual photos/shots, not for an automatic, large-scale generation of data. 
We do not necessarily need to generate high-quality visual data since what matters is the usefulness of the generated data for improving the person detector performance. 
There are several lines of work demonstrating improvements in various recognition tasks, even with non-photorealistic training data. The two prime examples are: (i) recent advances in domain randomization in robotics~\cite{Tobin2017Domain,Loing2019Virtual} where object detectors
are trained using synthetic scenes with random foreground
and background textures, and (ii) optical flow estimators
trained using the \emph{flying chairs} dataset~\cite{Dosovitskiy2015FlowNet}, which
pastes synthetically generated chairs onto a random background
images. We follow this line of work and aim primarily
at covering the different modes of appearance variation
(pose, clothing, background) rather than synthesizing
photorealistic outputs.

\paragraph{Generative data augmentation.}
We build on Generative Adversarial Networks (GANs)~\cite{Goodfellow-2016-Nips,WassersteinGAN,Lin-2018-PacGAN,salimans2016improved,Dumoulin-2016-Adversarially,Nowozin2016fGANTG,Xiao-2018-BourGAN,Donahue-2016-adversarial,Radford-2015-Unsupervised,Reed-2016-Generative,Metz-2016-Unrolled},
which have shown a great progress recently in generating visual data~\cite{CGAN,wang2018high,Isola2017Image,Zhu-ICCV-2017,liu2017unsupervised,huang2018multimodal,SPADE,KarrasALL18,karras2018style,shaham2019singan}. 
GANs have also been useful for various forms of data augmentation, including (i) adding synthetic samples to expand a training set of real examples~\cite{DwibediMH17}; (ii) improving the visual quality of synthetic samples generated via computer graphics~\cite{huang2017expecting}; or (iii) generating variants of original real samples~\cite{wang2018high,choi2018stargan,PumarolaAMSM18}.
Recent progress in conditional image generation has enabled the generation of plausible images, and videos of never seen before humans (full body and faces)~\cite{Wu_2019_ICCV,karras2018style}, but these methods generate full images and cannot insert people into given backgrounds. Others have considered re-animating (re-targeting, puppeteering) real people in real scenes~\cite{Chan2018Everybody,thies2016face2face,Ma2017Pose,Dong2018Soft,Balakrishnan2018Synthesizing}. These methods have demonstrated impressive visual results in image/video editing set-ups but are not geared towards large-scale generation of training data.
We build on this work and develop an approach for disentangled and controlled augmentation of training data in automotive settings, where in contrast to previous work, our approach produces a full scene by explicitly estimating the foreground mask and compositing the pedestrian(s) into the background scene.

\paragraph{Synthesizing people.}
Other related methods, aiming to synthesize images of people, typically use conditional GANs.
One line of work aims at changing the pose of a given person or changing the viewpoint. This is achieved either by using a two-stage approach consisting of pose integration and image refinement~\cite{Ma2017Pose}, by performing the warping of a given person image to a different pose with a specially designed GAN architecture~\cite{Dong2018Soft}, or by requiring segmentation of the conditional image into several foreground layers~\cite{Balakrishnan2018Synthesizing}. Some works consider different factors of variation in person generation~\cite{Ma2015Disentangled}, but do not insert people into given full scenes in automotive settings as we do in our work.
Others aim at changing only the pose~\cite{Balakrishnan2018Synthesizing,Liu_2019_ICCV,Siarohin2018Deformable} or viewpoint~\cite{Si2018Multistage} of a given person in a given image keeping the appearance (e.g., clothing) of the person and the background scene fixed. In contrast, our model is designed to control independently the person's pose, appearance and the background scene.

The recent work~\cite{Wu2019PMC} uses as an input a silhouette (mask) sampled from an existing database of poses, which limits the diversity of the generated outputs. Our approach uses a generative model for poses (represented as keypoints), and we estimate the silhouette automatically. This allows for more complete control and higher diversity of output poses, including rare ones. 

Others have also considered detailed 3D modeling of people and the input scene~\cite{Zanfir2020Human}, or modeling individual component attributes (e.g., mixing clothing items such as shorts and a tank top from different input photographs~\cite{Men2020Controllable}. While the outputs of these models are certainly impressive, their focus (and their evaluation) is on generating visually pleasing fashion photographs in, typically, indoor scenes with a limited variation of imaging conditions.  In contrast,  our model does not require 3D models of people or individual clothing attributes as input; we focus on demonstrating improved person detection performance in automotive settings and consider the appearance, pose, and the background scene in a single data generator model, which allows us to handle scenes with widely changing imaging conditions (e.g., day, night, snow).

The problem of enlarging the training dataset has been recently explored in~\cite{Liu_2019_ICCV} where the generative model and a detector are optimized jointly, and in~\cite{Wu_2019_ICCV} where a class-conditional GAN is used for synthesizing pedestrian instances. In contrast to these works, we train the generator and classifier separately and focus on having full control of generated person images as well as integration into complete scenes.

\paragraph{Cut-and-paste data augmentation methods.}
Cut-and-paste techniques can be also used to insert humans or objects into the given background image either randomly \cite{DwibediMH17}, ensuring a global visual consistency \cite{DBLP:journals/corr/GeorgakisMBK17,lee2018context}, or using 3D human models \cite{varol2017learning,chen2016synthesizing,pishchulin2017building,zhou2010parametric}. 
The automotive setting related to our work has been considered in~\cite{huang2017expecting}, emphasizing the importance of detecting corner cases such as people in rare poses, children playing on the street, skateboarders, etc. 
To this end, the authors have collected an annotated dataset of pedestrians in dangerous scenarios obtained from a computer game engine. Using straight-up computer graphics generated samples is an approach parallel to ours. While it may offer higher image quality, 
it struggles to capture the diversity of the real world (e.g., texture, clothing, backgrounds)~\cite{Hattori2020Synthesizing,Marin2010Learning} or require \PP{to control} a lot of precise scene parameters  
(geometry, environment maps, lighting) and some user intervention~\cite{Alhaija2018Augmented}.

%% file: include/method.tex
\section{\PP{Proposed architecture and loss}} \label{sec:method}

\begin{figure*}[tb]
     \centering
     \includegraphics[width=0.99\linewidth]{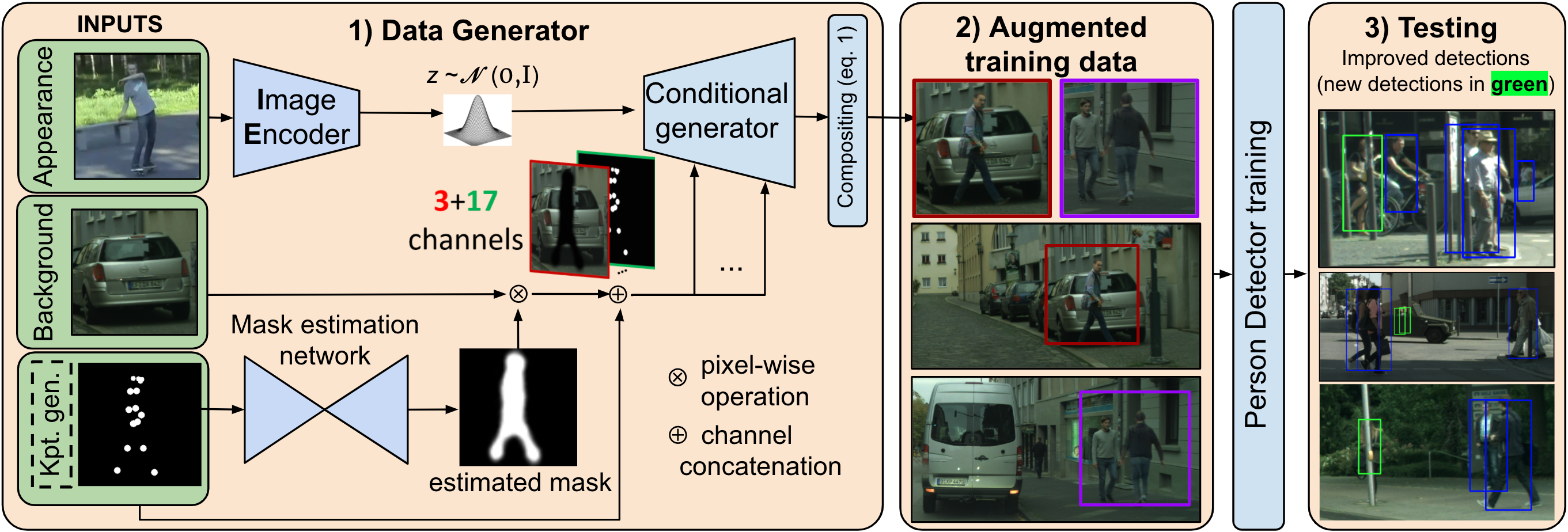}
\vspace{-1.5ex}
\caption{\small \textbf{Augmenting training data with our DummyNet approach.} 
The inputs are the desired pose (keypoints), desired pedestrian appearance (specified by an image), and the target background scene. The output is a scene with the composited pedestrian.
The DummyNet data generator (1) augments the training data (2), which leads to improved person detector performance (3). 
The data generator box (1) displays the inference-time setup. 
The training scheme and visualizations of the proposed new losses are available in the appendix.
}
\label{fig:training_scheme}
\vspace{-4ex}
\end{figure*}

Our objective is the complete control of the target person's \emph{pose}, \emph{appearance}, and \emph{background}.
For example, suppose it is difficult to collect large amounts of training data of people in the night or snowy conditions, but ``background'' imagery (with no or few people only) is available for these conditions. Our approach allows synthesizing new scenes that contain people with a variety of poses and appearances embedded in the night and snowy backgrounds.

To achieve that, we have developed \emph{DummyNet}, a new generative adversarial network architecture that takes the person's \emph{pose}, \emph{appearance} and \emph{target background} as input and generates the output scene containing the composited person. Our approach has the following three main innovations. First, the control is achieved by explicitly conditioning the generator on these three different inputs.
Second, the model automatically estimates a mask that separates the person in the foreground from the background.
The mask allows focusing compute on the foreground or background pixels as needed by the different components of the model \AV{and is also used in calculating the loss function}. 
Third, the architecture is trained with a new loss ensuring better appearance control of the inserted people. 
The overview of the entire approach is shown in Fig.~\ref{fig:training_scheme}.

The rest of this section describes the individual components of the model organized along with the three main areas of innovation. Additional details of the architecture, including the detailed diagrams of training and inference, as well as losses, are in the appendix.

\subsection{Controlling pose, appearance and background}
\label{sec:control}

As illustrated in Fig.~\ref{fig:training_scheme}, at inference time, the model takes three inputs (green boxes) that influence the {\em conditional generator}:  (i) person's appearance (clothing, hair, etc.) is specified by an image sampled from a dataset of training images containing people and encoded by an {\em image encoder}.  
(ii) background scene (typically an urban landscape) is sampled from a dataset of background scenes;
finally, (iii) person's keypoints are produced by a {\em keypoint generator}. 
The main components of the architecture are described next.

\vspace{-2ex}
\paragraph{Conditional generator.}
\label{sec:gen_topology}

The generator takes as input the latent person appearance code, the background scene (with masked-out foreground pixels), and target keypoints in the form of 17 individual-channel heatmaps, one for each keypoint. 
The keypoint representation allows for fine control over the person's pose, including the ability to distinguish between frontal/back as well as left/right views, which would not be possible when using a person mask only.
Conditioned on these inputs, the generator outputs an image of a person placed into the given background image in the pose given by the person keypoints and with the appearance specified by the provided appearance code (latent vector).
This latent vector is passed through a fully-connected layer and further to the convolutional layers of the generator.
The background image is concatenated with the keypoint heatmaps and used as an additional input to the generator. 
The generator architecture is based on residual blocks containing SPatially-Adaptive (DE)normalization layers~\cite{SPADE} and uses progressive growing. In each such block, the keypoints and the background are injected into the network through the SPADE layer.
The corresponding patch discriminator based on~\cite{Isola2017Image} is described in the appendix secion~\ref{sec:dis_topology}.

\vspace{-2ex}
\paragraph{Person appearance encoder.} We pre-train a variational autoencoder (VAE) separately from DummyNet.
During DummyNet training, we use the encoder part of this VAE as a person appearance encoder. Its architecture comprises convolutional layers that take as input a $64\times64$ px person image with masked-out background and produce a latent vector encoding the appearance. 

\vspace{-2ex}
\paragraph{Keypoint Generator.} \label{sec:dis_topology}
The keypoint generator was created from the OpenPose~\cite{Cao2018OpenPose} training set via viewpoint and pose clustering followed by principal component analysis within each pose cluster. This set of simple models fully captures the pose distribution and allows us to sample from it and use the result as input to the generator and the mask estimation network.

\subsection{Mask estimation and fg/bg compositing}

An essential component of our approach is the {\em mask estimator (ME)} that predicts the foreground/background (fg/bg) segmentation mask from the input set of keypoints.  
This is a U-Net-type network~\cite{Ronneberger2015UNet}
that ingests keypoint locations encoded in $17$ channels (\AV{COCO format}, one channel per keypoint), and outputs a mask predicting which pixels belong to the person. 
 This output is a one-channel map $\mathcal{M} \in [0, 1]^{H\times W}$, where $(H,W)$ are the height and the width of the output image, respectively. Output mask values are in the range of $[0, 1]$ to composite the generated person image smoothly into the target real background image. We pre-train this network separately from DummyNet using person mask and keypoint annotations in MS COCO dataset~\cite{lin2014coco}.
Using the pre-trained mask estimator network $\mathrm{ME}$, we obtain the output mask $\mathcal{M}$ from the given input skeleton keypoints as $    \mathcal{M} = \mathrm{ME} \left( \texttt{kpts} \right)$.
The estimated fg/bg mask is used at different places of the architecture as described next.

First, it is used to produce the foreground person image \DHh{as input to the encoder producing the appearance vector} and the background scene \DHh{with masked out pixels prepared for foreground synthesis} that are used as inputs to the conditional generator. Details are in the learning and inference diagrams in the appendix. 

Second, the fg/bg mask is used to composite the final output. Given a background image $\texttt{I}_{\text{bg}}$ and the output of the conditional generator $\texttt{I}_\text{gen}$, the final \PP{augmented training} image $\texttt{I}_\text{aug}$ is obtained by compositing the generated foreground image with the background scene as 
\begin{equation}
    \texttt{I}_\text{aug}  = \mathcal{M} \odot \texttt{I}_\text{gen} + \left( 1-\mathcal{M} \right) \odot \texttt{I}_\text{bg}, \label{eq:OUT}
\end{equation}
where $\odot$ denotes the element-wise multiplication and $\mathcal{M}$ the estimated mask.
In contrast to other works, we do not need to use manually specified masks during the inference time as we estimate them automatically with the Mask Estimator. This allows us to process more data and hence a more diverse set of poses.
If the manual mask is available, it can be used as well.
Finally, the estimated person mask is also used to compute fg/bg appearance losses as described next. 

\subsection{Masked losses and training}
\label{sec:losses}

At training time, the model takes as input an image that is separated into the foreground person image and the background scene using the estimated fg/bg mask.
The model is then trained to produce a realistic-looking output that locally fits the background scene and uses the appearance conditioning from the input foreground image. The network is trained with a weighted combination of four losses, described next.

For the discriminator, we use the Improved Wasserstein loss (WGAN-GP)~\cite{Gulrajani-NIPS-2017} as we found it more stable than other adversarial losses. The loss measures how easy it is to discriminate the output image composite given by Equation~\ref{eq:OUT} from the real input training image. 

Next, we introduce a new loss term that we found important for inserting pedestrians into a variety of backgrounds. 
This \AV{person appearance consistency loss} $\mathcal{L}_\textrm{app}$ encourages the generator to change the appearance of the inserted pedestrian with the change of the input person appearance latent vector. This is achieved by enforcing the latent appearance vector of the output image to match the latent appearance vector provided at the input of the generation process, measured by $L_1$ distance.
This is implemented as \begin{align}
    \mathcal{L}_\textrm{app} &= \left\| \mathrm{ENC}  \left(\mathcal{M} \odot \texttt{I}_\text{in} \right) - \mathrm{ENC} \left(\mathcal{M} \odot \texttt{I}_\text{gen} \right) \right\| _1,
    \label{eq:appearance}
\end{align}
where $\mathrm{ENC}$ is the Image Encoder,
$\mathcal{M}$ is the estimated person mask,  $\texttt{I}_\text{in}$ is the input image, and $\texttt{I}_\text{gen}$ is the generated output. Please note how the estimated foreground mask $\mathcal{M}$ allows focusing the person appearance consistency loss on the person foreground pixels. An experiment showing the effect of appearance preconditioning can be found in the appendix.

We further add two reconstruction losses to our objective function, but have them act only on the foreground person pixels via the estimated person mask.
The first reconstruction loss, $\mathcal{L}_\textrm{Rec-dis}$, compares features extracted using the discriminator from the real input and generated output samples. The second reconstruction loss, $\mathcal{L}_\textrm{Rec-VGG}$ compares the real input and generated output in the feature space of a VGG19 network~\cite{VGG11} pre-trained for ImageNet classification. In both cases, the losses are modulated by the foreground mask to focus mainly on the foreground person pixels, similar to~\eqref{eq:appearance}, and use $L_1$ distance. We found these losses greatly stabilize training. 
The final loss is a weighted sum of the terms described above: \begin{equation}
\mathcal{L} = \lambda_1 \mathcal{L}_\textrm{WGAN-GP} + \lambda_2 \mathcal{L}_\textrm{Rec-dis} + \lambda_3 \mathcal{L}_\textrm{Rec-VGG} + \lambda_4 \mathcal{L}_\textrm{app}, 
\label{eqn:overal_loss}
\end{equation}
where hyperparameters $\lambda_i$'s are set experimentally to ensure convergence and prevent overfitting.
\DHh{Please see the appendix for ablations measuring (among others) the influence of our proposed appearance loss and the importance of having control over the background scene as the input to the generator.}

%% file: include/experiments.tex
\section{Experiments} 
\label{sec:experiments}

\begin{figure*}[ht!]
\centering
\begin{subfigure}{0.99\linewidth}
\centering
\includegraphics[width=0.333\linewidth]{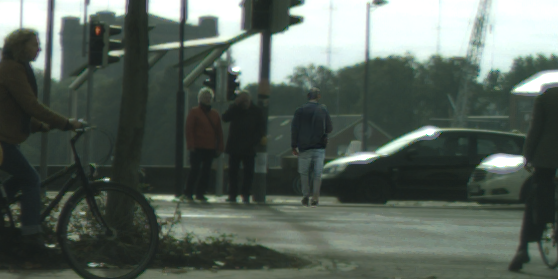}%
\includegraphics[width=0.333\linewidth]{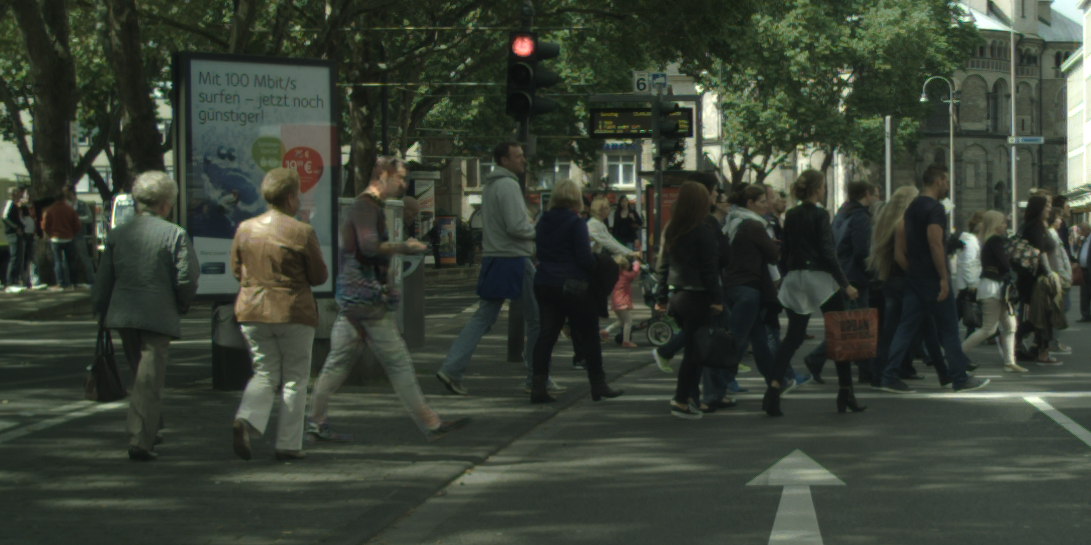}%
\includegraphics[width=0.333\linewidth]{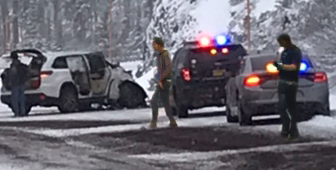}
\end{subfigure}
\begin{subfigure}{0.99\linewidth}
\centering
\includegraphics[width=0.167\linewidth]{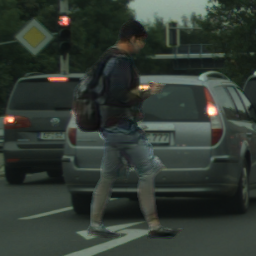}%
\includegraphics[width=0.167\linewidth]{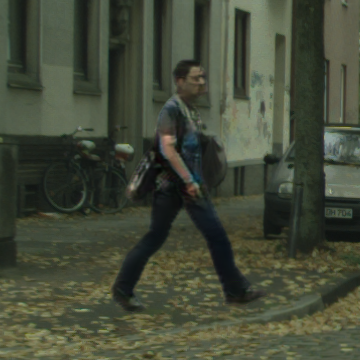}%
\includegraphics[width=0.167\linewidth]{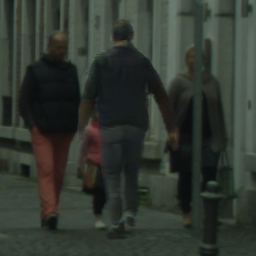}%
\includegraphics[width=0.167\linewidth]{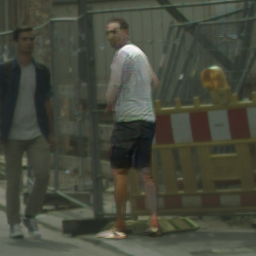}%
\includegraphics[width=0.167\linewidth]{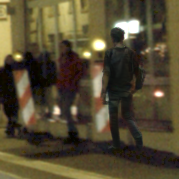}%
\includegraphics[width=0.167\linewidth]{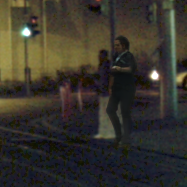}
\end{subfigure}
\vspace{-1ex}
\caption{{\bf Artificial people in urban scenes produced by our DummyNet approach.} Examples are shown at various scales and levels of detail. \emph{Top}: Each full-scene image contains one synthetic person with the exception of the winter scene where there are two.
\emph{Bottom}: Close-ups of day and night scenes with one synthetic person in each.
{\bf Please see the appendix for additional qualitative results as well as ablations evaluating the importance of the different contributions.}}
\label{fig:in_the_wild}
\vspace{-3.5ex}
\end{figure*}

In this section, we present a series of experiments on controlled dataset augmentation with the aim to improve the accuracy of a person classifier/detector in the context of autonomous driving. The augmentation is controlled as our DummyNet can generate images of people with a specific distribution of poses, appearances, and backgrounds. We consider four experimental set-ups. 
The first experiment (Sec.\,\ref{sec:exp1}) focuses on augmenting the daytime Cityscapes dataset. We test data augmentation in the low-data regime, i.e., with insufficient real training data for training the person classifier. 
In the second experiment (Sec.\,\ref{sec:night}), we use DummyNet to generate night-time person images and show significant improvements in classifier performance on the NightOwls dataset~\cite{Neumann2018NightOwls}.

In the next experiment (Sec.\,\ref{sec:experiments_csp}) we use DummyNet to improve performance of the  state-of-the-art person detection network CSP~\cite{liu2019high} in the standard (full) data regime on the Cityscapes \AV{and Caltech} datasets.
\AV{Finally, we demonstrate the benefits of our approach in set-ups (Sec.\,\ref{sec:experiments_nightowls_det}) where we have only access for training to day-time annotated images (CityPersons) along with night-time images devoid of people, and we wish to detect pedestrians at night (NightOwls). } \AV{See Fig.~\ref{fig:in_the_wild} for images generated by DummyNet.}

\vspace{-3ex}
\paragraph{Person classifier.}
In experiments~\ref{sec:exp1} and~\ref{sec:night}, we consider a CNN-based classifier with $6,729$ parameters, which is a realistic set-up for a digital signal processor in a car (where GPU is not available in a deployed system) and is trained \emph{from scratch}. The classifier consists of $4$ convolutional layers with a $3\times3$ kernel, stride $2$, ReLU activations, maxpooling, and one fully connected layer with sigmoid activation. For both experiments, the classifier is trained for $1,000$ epochs and the one with best validation error is kept. 

\vspace{-3ex}
\paragraph{Training data for DummyNet.}
For training the generator, we aim for the real-world data with all its artifacts (blur, low/high res, changing lighting, etc.) and with enough samples with person height of at least $190$px. 
To achieve that, we leveraged the YoutubeBB~\cite{RealSMPV17} dataset that contains a large number of videos with people in a large variety of poses, illuminations, resolutions, etc.
We used OpenPose~\cite{Cao2018OpenPose}
to automatically detect people and annotate up to $17$ keypoints 
based on their visibility. 

The final dataset contains $769,176$ frames with annotated keypoints.
Please note that the keypoints and mask annotations (estimated by our mask estimator) are noisy as they have been obtained automatically, yet, are sufficient for our needs. More details about the dataset, examples of training images with skeleton annotations, and the keypoint generator are in the appendix.

\subsection{Data augmentation in a low-data regime} \label{sec:exp1}

In this experiment, we show how adding training person samples generated by DummyNet influences the testing performance of the person classifier.
We conduct an experiment where there is only a small number of training samples available and investigate how adding synthesized images of the positive class (person) helps the performance of the resulting classifier.
We compare against two baseline methods \emph{Cut, Paste and Learn} (CPL)~\cite{DwibediMH17} and pix2pixHD~\cite{wang2018high}. Both methods require stronger input information as they need to have a segmentation mask of the inserted object. In addition, pix2pixHD requires to have a complete segmentation of the input scene. Therefore pix2pixHD and CPL have an advantage of additional information that our DummyNet does not require.

In Table~\ref{tab:exp1}, we report results in the form of a miss rate at $1\%$ and $10\%$ false positive rate. We compare results to the baseline classifier trained with only 100 real and no synthetic samples (first row) and investigate adding up to 1000 synthetic samples. 
In this low-data regime, the classifier trained on real data only performs poorly, and adding synthesized images clearly helps. However, when too many synthetic examples are added, performance may decrease, suggesting that there is a certain amount of domain gap between real and synthetic data.  
Compared to the pix2pixHD and CPL, our DummyNet performs the best, bringing the largest performance boost, lowering the baseline miss rate by $21.7\%$ at $1\%$ FPR and by $25.7\%$ at $10\%$ FPR.

\begin{table}
\setlength{\tabcolsep}{3.5pt}
\centering 
\footnotesize
 \begin{tabular}{c|cc|cc|cc}
 \toprule

generated & \multicolumn{2}{c|}{pix2pixHD} & \multicolumn{2}{c|}{CPL} & \multicolumn{2}{c}{DummyNet} \\
  \midrule
FPR & $1\%$ \em & $10\%$ & $1\%$ & $10\%$ & $1\%$ & $10\%$  \\ 
  \midrule
 $0$ & $0.980$ & $0.682$ & $0.980$ & $0.682$ & $0.980$ & $0.682$ \\
 $75$ & $0.852$ & $0.424$ & $\mathbf{0.795}$ & $\mathbf{0.467}$ & $\mathbf{0.763}$ & $\mathbf{0.425}$ \\
 $200$ & $\mathbf{0.800}$ & $\mathbf{0.438}$ & $0.809$ & $0.466$ & $0.836$ & $0.490$ \\
 $500$ & $0.865$ & $0.616$ & $0.818$ & $0.567$ & $0.861$ & $0.636$ \\
 $1000$ & $0.922$ & $0.731$ & $0.813$ & $0.514$ & $0.790$ & $0.463$ \\
\bottomrule 
\end{tabular}
\vspace{-2ex}
\caption{\small 
	{\bf Data augmentation in low-data regimes on the Cityscapes dataset.} We report classifier test set miss rate (lower is better). 100 real samples were used for training, plus different amounts of generated samples (leftmost column) for augmentation. Test results are reported at $1\%$ and $10\%$ false positive rate (FPR). The best results are marked for $1\%$ FPR in bold for each method.}
\label{tab:exp1}
\vspace{-4ex}
\end{table}

%% file: include/experiment_nightowls.tex
\subsection{Person classification at night time} \label{sec:night}

Annotated training images with pedestrians at night are hard to obtain. On the other hand, it is relatively easy to get night scenes that contain no people. To this end, we construct an experiment on the NightOwls dataset~\cite{Neumann2018NightOwls} and vary the number of available real (night-time) training person images. We then add more images of persons at night synthesized by DummyNet. Generated samples were obtained by providing the generator with day-time (but low light, based on thresholding the average brightness) input images of people together with night-time background scenes to get night-time output scenes with people. For the complete setup, please see the appendix.

Classification results on \emph{testing NightOwls data} are shown in Table~\ref{tab:exp_nightowls} and demonstrate that generated samples help to train a better classifier, which improves over the baseline by a large margin. The reported results are mean miss rates over five runs. In low-data regime, we can lower the miss rates by nearly $20\%$. Adding synthetic examples improves performance in all settings, even with an increasing amount of real training data.
In particular, note that we can improve performance even when having all the available real training data (column `full set'). In that case, we improve MR at $10\%$ FPR by more than $6\%$. Please note that DummyNet was not finetuned on this dataset. Please see additional results and examples of synthesized night-time data in Fig.~\ref{fig:in_the_wild} and the appendix.

\begin{table}[t]
\setlength{\tabcolsep}{4pt}
\centering 
\footnotesize
 \begin{tabular}{c|cc|cc|cc|cc}
 \toprule

gen\textbackslash real  & \multicolumn{2}{c|}{$0$} & \multicolumn{2}{c|}{$100$} & \multicolumn{2}{c|}{$1000$} & \multicolumn{2}{c}{$12000$ (full)} \\
  \midrule
FPR & $1\%$ \em & $10\%$ & $1\%$ & $10\%$ & $1\%$ & $10\%$ & $1\%$ & $10\%$  \\ 
  \midrule
 0 & & & 0.88 & 0.62 & 0.64 & 0.34 & 0.51 & 0.28 \\
 5k & 0.76 & 0.49 & 0.72 & 0.44 & 0.63 & \bf{0.33} & 0.48 & 0.24 \\
 10k & \bf{0.71} & \bf{0.46} & 0.72 & \bf{0.42} & \bf{0.58} & \bf{0.33} & \bf{0.47} & 0.25 \\
 20k & 0.76 & 0.52 & \bf{0.72} & 0.42 & 0.62 & 0.36 & 0.50 & \bf{0.22} \\
\bottomrule 
\end{tabular}
\vspace{-2ex}
\caption{\small 
	{\bf Night-time person detection on the NightOwls dataset.} We report the mean classifier test set miss rate (lower is better) over 5 runs. Test results are reported at $1\%$ and $10\%$ FPR. The best results for every combination are shown in bold. Samples generated by our method help to train a better classifier, often by a large margin, over the baseline trained only from real images (the first row corresponding to 0 generated samples). 
}
\label{tab:exp_nightowls}
\vspace{-4ex}
\end{table}

%% file: include/experiment_csp.tex
\subsection{Improving state-of-the-art person detector} \label{sec:experiments_csp}

\textbf{Citypersons.} We conduct an experiment with one of the state-of-the-art person detectors, CSP~\cite{liu2019high}, on the full Citypersons dataset~\cite{zhang2017citypersons}.
We use our DummyNet to extend the training set of this dataset. The augmentation procedure is the same for all the compared methods. It uses the semantic segmentation of the road scenes to place pedestrians at plausible locations (ground, road, sidewalk) and uses existing people in the scene (if available) to choose the position and size of the inserted people. We require that the new person stands on the ground and does not occlude other people already present in the image.  The details of the augmentation procedure are given in the appendix.  

Following~\cite{liu2019high}, we train the detection network for 150 epochs and report log-average miss rate for multiple setups of the detector with the best validation performance, see Table~\ref{tab:exp_csp}. We compare the following data augmentation setups: (a) original results reported in~\cite{liu2019high}; (b) our CSP detector retrained on original Citypersons images to reproduce results of~\cite{liu2019high};  
(c) augmentation with the SURREAL~\cite{varol2017learning} dataset, (d) CPL augmentation~\cite{DwibediMH17}; 
\AV{(e) augmentation with recent ADGAN~\cite{Men2020Controllable} generative network;}
(f)-(g) augmentation with the Human3.6M dataset~\cite{h36m_pami,IonescuSminchisescu11} using provided segmentation or segmentation with DeepLabv3+~\cite{chen2018encoder};
and (h) training on DummyNet extended Cityscapes dataset.

We observe a consistent improvement in performance across all setups (see appendix for more details) when DummyNet augmented samples are used. Please note that differences between the reported results in~\cite{liu2019high} (a) and our reproduced results (b) could be attributed to differences in initialization (random seed not provided in~\cite{liu2019high}); otherwise we use the same training setup.

\textbf{Caltech.}
Following the experiments and the setup in~\cite{liu2019high}, we also train a CSP detector on the Caltech~\cite{dollar2011pedestrian} dataset. We \AV{initialize the detector weights} with the best-performing network on CityPersons. Our data augmentation improves over the results reported in~\cite{liu2019high}  (the reasonable setup), reducing the LAMR from $3.8\%$ to $3.47\%$, i.e., by almost $0.35\%$,  which is non-negligible given the overall low LAMR. These results demonstrate the benefits of our approach on another challenging dataset.

\begin{table}[th]
\centering 
\footnotesize
\begin{tabular}{p{3.35cm} | p{1.25cm} | p{1.05cm} | p{1.cm} }
\toprule 

 setup & reasonable & small & partial \\
 \midrule
 (a) CSP reported & $11.02\%$ & $15.96\%$ & $10.43\%$ \\
 (b) CSP reproduced & $11.44\%$ & $15.88\%$ & $10.72\%$ \\
 (c) SURREAL aug. & $11.38\%$ & $17.39\%$ & $10.56\%$ \\
 (d) CPL aug. & $11.36\%$ & $16.46\%$ & $10.84\%$ \\
 (e) ADGAN aug. & $10.85\%$ & $16.20\%$ & $10.55\%$ \\
 (f) H3.6M aug., orig. & $11.07\%$ & $16.66\%$ & $10.58\%$ \\
 (g) H3.6M aug., DLv3 & $10.59\%$ & $16.00\%$ & $10.21\%$ \\
 (h) DummyNet aug. (ours) &  $\mathbf{10.25\%}$ &  $\mathbf{15.44\%}$ & $\mathbf{9.12\%}$ \\
\bottomrule 
\end{tabular}
\vspace{-2ex}
\caption{\small \textbf{Improving state-of-the-art person detector~\cite{liu2019high}.} Log-average miss rate of the detector (lower is better) in multiple testing setups (reasonable, small, partial) provided in~\cite{liu2019high}.}
\label{tab:exp_csp}
\vspace{-4ex}
\end{table}

%% file: include/nightowls_detection.tex
\subsection{Person detection in night-time scenes} \label{sec:experiments_nightowls_det}
Here we address the problem of an insufficient amount of annotated training data again. We conduct an experiment where we have access only to annotated images captured during daytime (CityPersons dataset), and we wish to detect pedestrians at night time (NightOwls dataset). However, we do not have any nigh-time images annotated with people. \AV{This is a similar setup as in Section~\ref{sec:night}, but here we consider a well-known object detection architecture.}

As a baseline, we train a Faster-RCNN with ResNet-50 backbone initialized from COCO detection task with the day annotations only (setup (a)).
Then, we use the same augmentation and person placement strategy as described in Section\,\ref{sec:experiments_csp}, and retrain 
the Faster-RCNN detector using the augmented training dataset. The results are summarized in Table~\ref{tab:nightowls_det} and show \AV{that our method (f) performs the best, outperforming the baseline~(a) by $16\%$, the-state-of-the-art person generative network~(b) by $\sim12\%$, the compositing approach~(d) by $\sim8\%$, and the nearest competitor~(e), which uses much more images depicting people in various poses but with limited appearance variation, by $\sim3\%$ measured by the LAMR (reasonable setup). These results indicate that it is important to have the variability and the control over (i) the appearance, (ii) the pose, and (iii) the background scene of the generated people for person detection in automotive scenarios. The complete set} of the quantitative results is in the appendix.

\begin{table}[!ht]
\centering 
\footnotesize
\begin{tabular}{p{3.15cm} | p{1.25cm} | p{1.1cm} | p{1.1cm} }
\toprule 

 setup & reasonable & small & occluded  \\
 \midrule
 (a) CityPersons ann. & $41.90\%$ & $50.55\%$ & $65.95\%$ \\
 (b) ADGAN [Men20] & $36.60\%$ & $48.91\%$ & $54.49\%$   \\
 (c) SURREAL [Varol17] & $32.94\%$ & $44.05\%$ & $49.24\%$  \\
 (d) CPL [Dwibedi17] & $30.73\%$ & $49.61\%$ & $54.60\%$  \\
 (e) H3.6M [Ionescu14] & $27.83\%$ & $43.81\%$ & $45.67\%$ \\
 (f) DummyNet (ours) & $\mathbf{24.95\%}$ & $\mathbf{39.73\%}$ & $\mathbf{44.89\%}$ \\
\bottomrule 
\end{tabular}
\vspace{-2ex}
\caption{\small \textbf{Person detection in night-time scenes.} LAMR (lower is better) in multiple testing setups on the NightOwls dataset~\cite{Neumann2018NightOwls}.}
\label{tab:nightowls_det}
    \vspace*{-5ex}
\end{table}

%% file: include/conclusion.tex
\section{Conclusion} \label{sec:conclusion}

We have developed an approach for controlled augmentation of person image datasets, where people with adjustable pose and appearance can be synthesized into real urban backgrounds. We have demonstrated that adding such generated data improves person classification and detection performance, especially in low-data regimes and in challenging conditions  (like the night-time person detection) when positive training samples are not easy to collect. We have shown that neural networks of various model complexities designed for multiple tasks benefit from artificially generated samples, especially when we have control over the data distributions. These results open up the possibility to control many more parameters (e.g., weather conditions, age, gender, etc.) and make a step towards controllable training and validation process where generated dummies cover challenging corner cases that are hard to collect in real-world situations.

%% file: include/acknowledgements.tex
{\small
\noindent
\textbf{Acknowledgements}
This work was supported by Valeo, the Grant Agency of the Czech Technical University in Prague, grant No. SGS18/205/OHK3/3T/37, and the European Regional Development Fund under the project IMPACT (reg. no. CZ.02.1.01/0.0/0.0/15\_003/0000468).
}

%% file: supplementary_raw.tex
\newif\ifaaai\aaaitrue
In this supplementary material, we first describe in detail the training and inference phases of our DummyNet approach~(Section~\ref{sec:supp_method}). Next, we provide more details about our person appearance encoder~(Section~\ref{sec:encoder}) and our conditional generator~(Section~\ref{sec:conditional_generator}). In Section~\ref{sec:ablations}, we show ablations of the different parts of our network \AV{and losses}. Then, in Section~\ref{sec:supp_clust}, we provide an additional description of DummyNet training and testing sets as well as the description of the pose clustering method and the pose keypoint generator model.
In Section~\ref{sec:cs_augmentation}, we describe how we insert new person instances into existing scenes and show examples of such augmented images.
\AV{Finally, in section~\ref{sec:exp_details} we give further details of the experiments from the main paper, provide the reader with a full set of results, and show more examples of real/augmented images.}

\section{DummyNet training and inference} \label{sec:supp_method}

\begin{figure*}[th]
     \centering
     \includegraphics[width=1.0\linewidth]{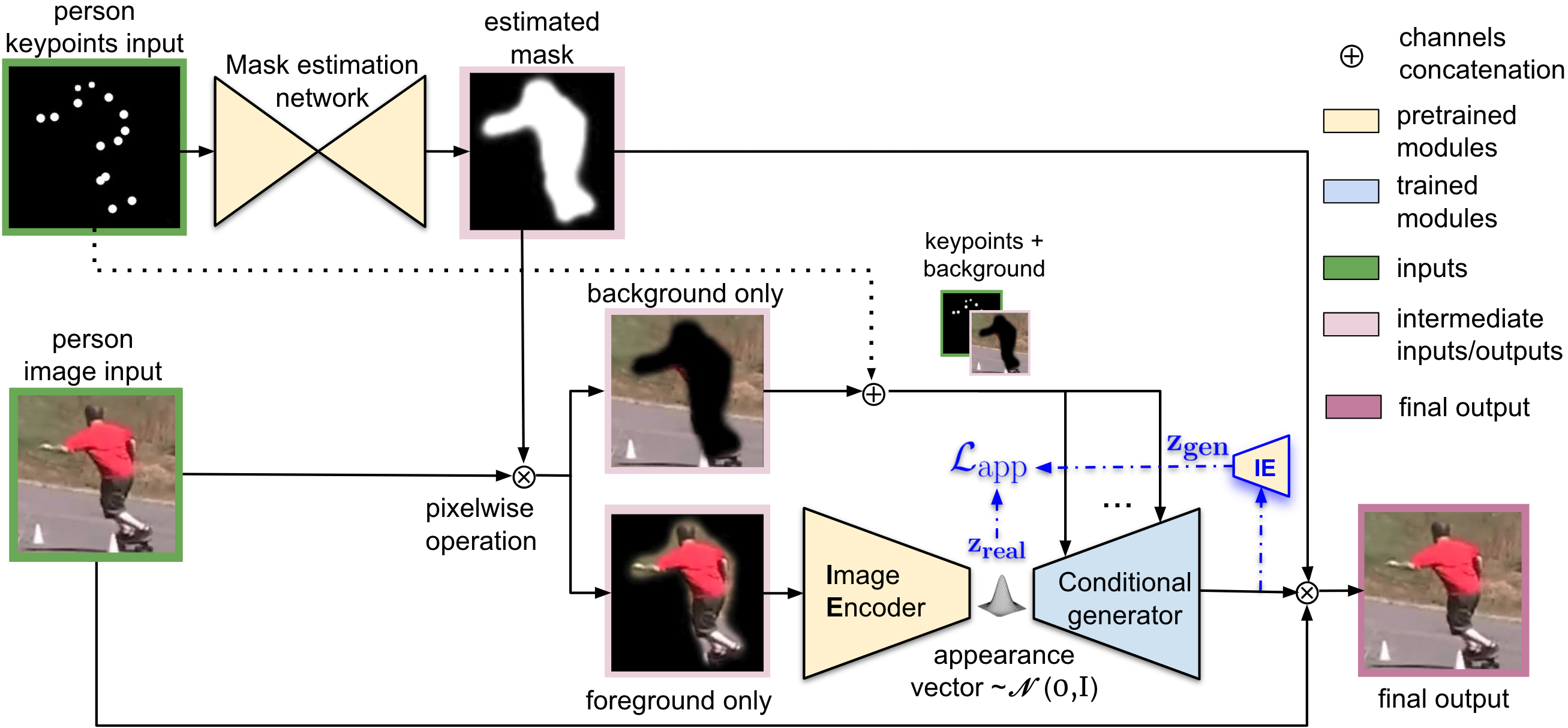}
\caption{\textbf{DummyNet training}. Green rectangles mark training inputs. The Image Encoder (IE) and Mask Estimation (ME) networks are pre-trained and kept fixed, while the conditional generator is trained. The newly proposed \emph{appearance loss} is highlighted in blue. Details of the loss computation are shown in Figure~\ref{fig:losses}.}
\label{fig:supp_scheme_train}
\end{figure*}

\begin{figure}[ht]
  \centering
  \includegraphics[width=1.0\linewidth]{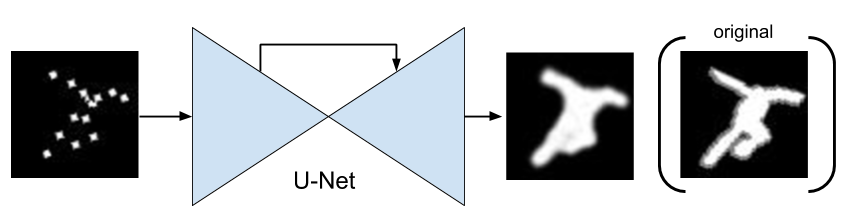}
  \vspace{-10pt}
  \caption{Input keypoints obtained by OpenPose or sampled by our keypoint generator are passed through the U-Net architecture to generate the output mask used for blending. Manually specified (``original'') mask is only used as a target during the Mask Estimator pre-training.}
\label{fig:mask}
\end{figure}

\paragraph{Training the generator.}  Figure\,\ref{fig:supp_scheme_train} shows the full diagram of generator training. Inputs are \emph{images of people} and the corresponding \emph{keypoints} detected by OpenPose~\cite{Cao2018OpenPose}. The mask estimation network (ME) is pre-trained on MS-COCO~\cite{lin2014coco} keypoints and masks (see Fig.~\ref{fig:mask}). The image encoder is pre-trained on MS-COCO images with foreground/background segmentation where the foreground (person) pixels are kept, and the background is removed using the mask. The images of people used for training the generator are obtained from the YoutubeBB dataset~\cite{RealSMPV17} with automatically estimated keypoints (OpenPose) and estimated masks (our ME).

\begin{figure*}[t]
        \centering
        \begin{subfigure}[b]{0.58\textwidth}  
            \centering           \includegraphics[height=3.3cm]{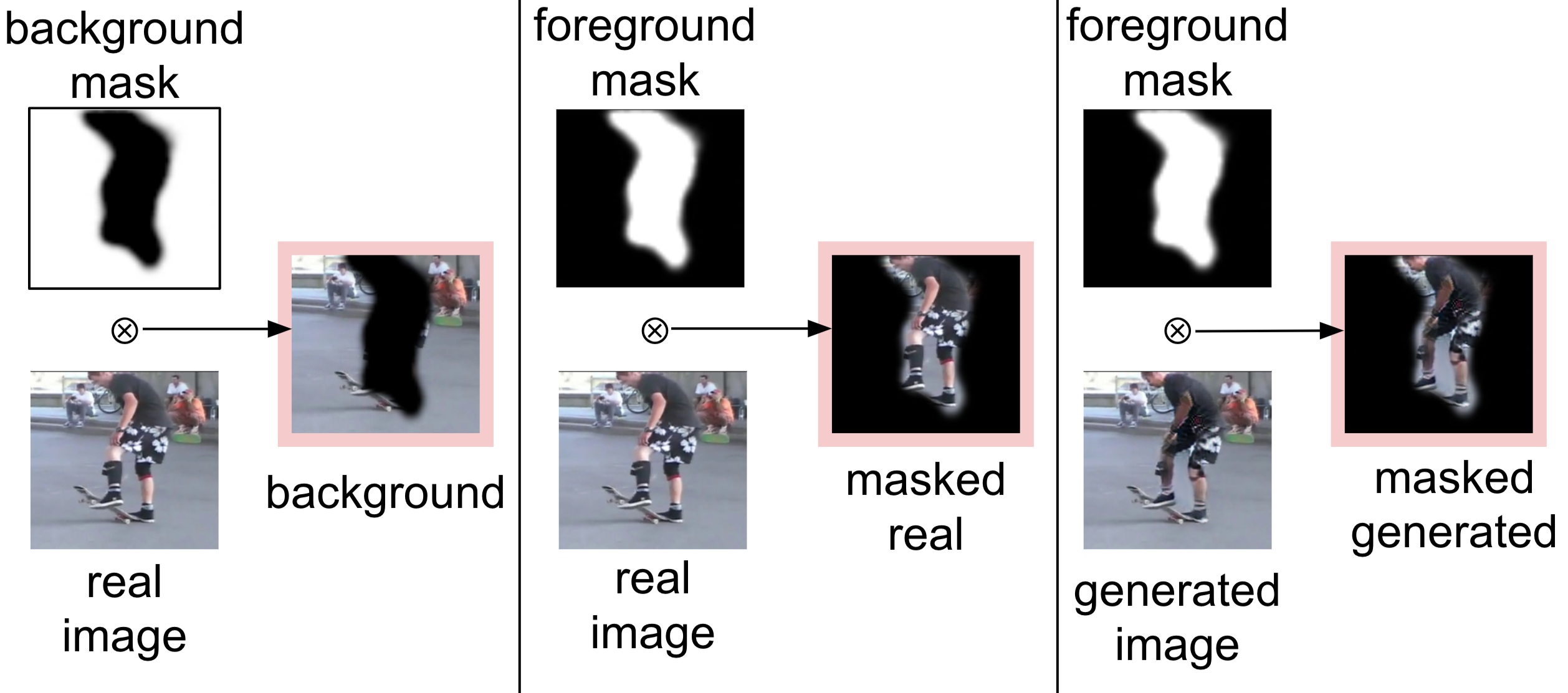}
            \caption[app]%
            {{\small Masking inputs for computing losses.}}    
            \label{fig:masking_out}
        \end{subfigure}
        \begin{subfigure}[b]{0.35\textwidth}   
            \centering             \includegraphics[height=3.4cm]{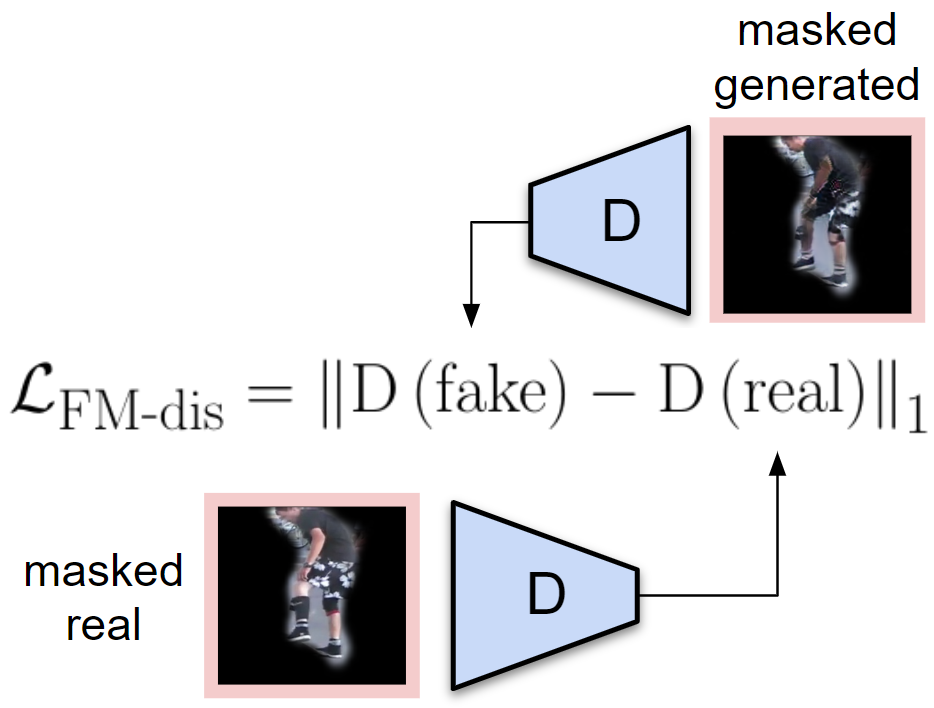}
            \caption[fmdis]%
            {{\small Discriminator feature-matching loss.}}    
            \label{fig:fmdis_loss}
        \end{subfigure}
        \vskip\baselineskip
        \begin{subfigure}[b]{0.48\textwidth}  
            \centering             \includegraphics[height=3.4cm]{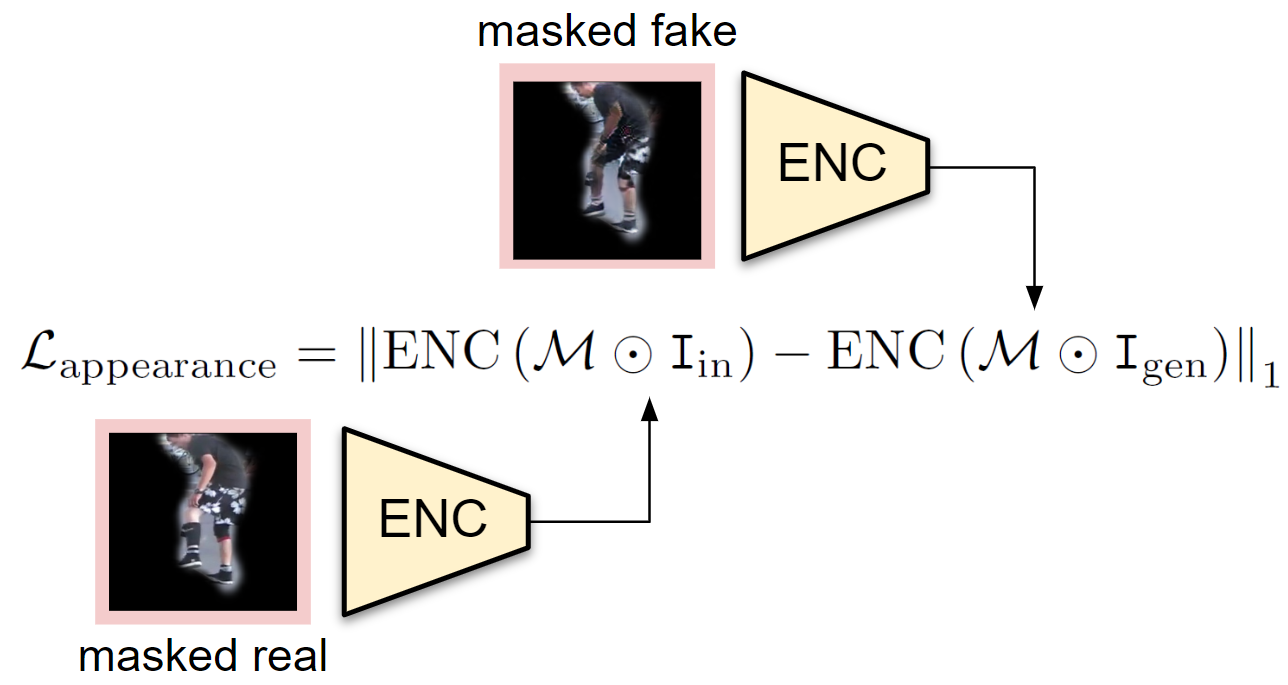}
            \caption[app]%
            {{\small Appearance loss.}}    
            \label{fig:app_loss}
        \end{subfigure}
        \begin{subfigure}[b]{0.48\textwidth}   
            \centering 
            \includegraphics[height=3.5cm]{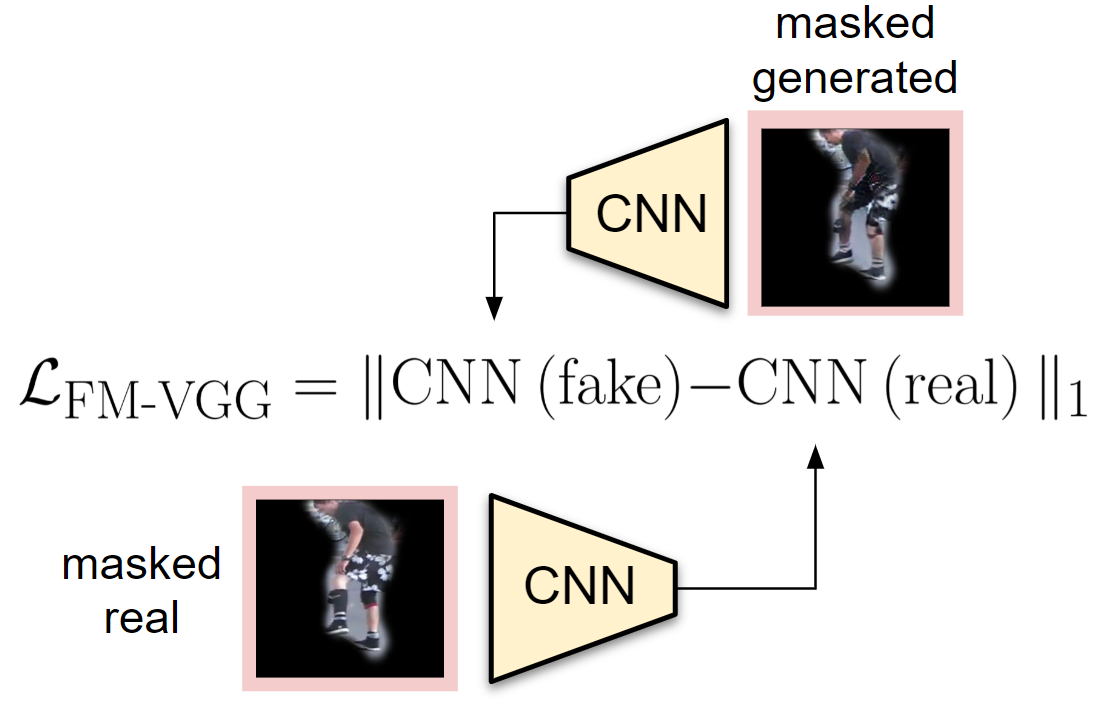}
            \caption[vgg_loss]%
            {{\small CNN feature-matching loss.}}    
            \label{fig:vgg_loss}
        \end{subfigure}
    \caption{\small {\bf Losses used in DummyNet training.}
    (a) \textit{Masking real and generated images for DummyNet loss}. The foreground mask is obtained by the Mask Estimator. We compute the background mask as the inversion of the foreground mask;
    (c) \textit{Appearance loss}: $\ell_1$ distance between appearance vectors of the real and the generated image; (b,d) \textit{feature matching losses}: $\ell_1$ distance between features (from trained discriminator or from pre-trained VGG network) of the real and the generated images.  
    The only trained network is the discriminator D. Both the CNN in the feature-matching loss~\cite{VGG11} and Appearance Encoder are pre-trained and fixed. We also use the standard WGAN-GP loss (not shown in this figure).} 
    \label{fig:losses}
\end{figure*}

The $17$ channels (one per each keypoint) are concatenated with the background image, where person pixels are masked out and used as input at multiple resolutions of the conditional generator.
In Figure~\ref{fig:masking_out}, we show how we apply the mask to real and generated images when computing the loss. Additional details of all the used losses are given in Figure~\ref{fig:losses}.

\begin{figure*}[th]
     \centering
     \includegraphics[width=\linewidth]{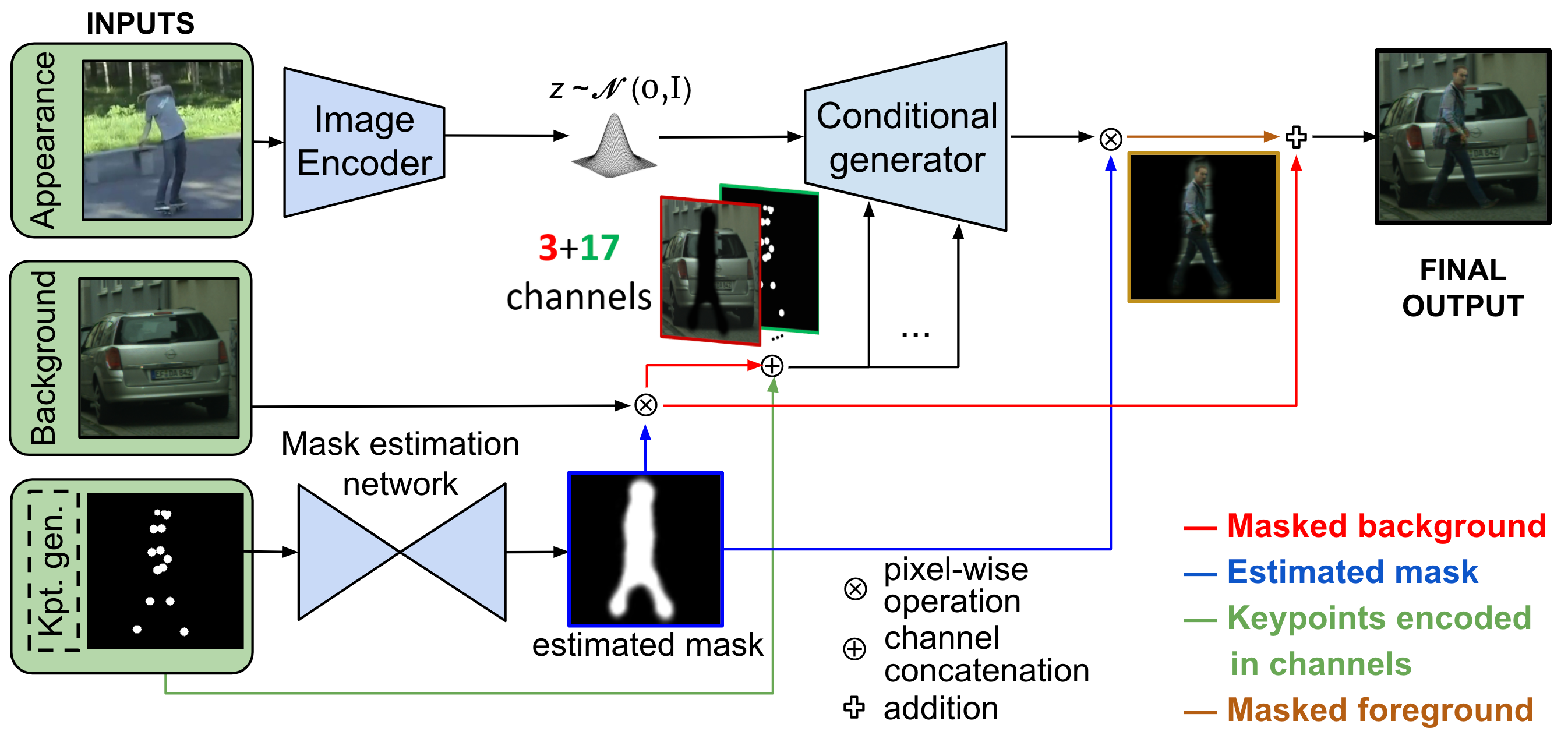}
\caption{{\bf Diagram of the data augmentation procedure (inference).} Green rectangles mark inputs to the generator. They include an automatic keypoint generator (PCA model from training skeletons keypoints). The appearance may be sampled from latent input or given by an existing person image.}
\label{fig:supp_scheme_infer}
\end{figure*}

\paragraph{Inference using our generator.} Figure\,\ref{fig:supp_scheme_infer} shows a more detailed version of Figure\,2 from the main paper. It is a full diagram of the person generation procedure. The procedure has two main components.  
The first main component is the person appearance encoder. As an input, it takes an image with a person, keypoints provided by OpenPose, and the mask estimated by our mask estimator. It outputs the person appearance latent vector that is used to condition the appearance of the generated person (for which the generator is trained using the appearance loss, as described later). The latent vector could be also sampled from a Gaussian distribution.
The second main component is the conditional generator. The generator takes the latent appearance vector, generated person keypoints, estimated mask, and the background image as input, and outputs the synthetic image of the person embedded in the background scene. The two main components, the appearance encoder and the conditional generator, are described in the next two sections.

\section{Person appearance encoder}
\label{sec:encoder}
\begin{figure*}[t!]
    \centering
    \begin{subfigure}[t]{0.4\textwidth}
        \centering
        \includegraphics[width=\textwidth]{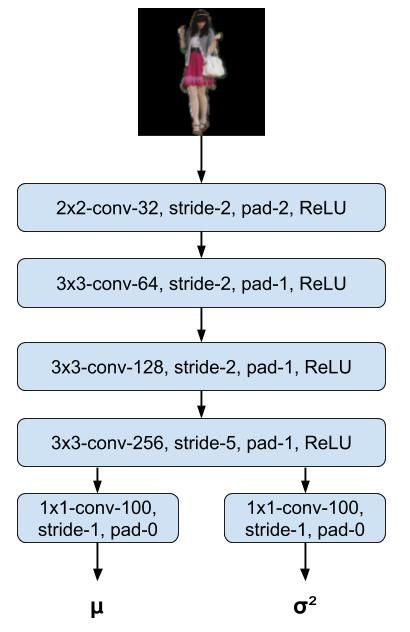}
        \caption{Encoder topology.}
        \label{fig:encoder_topology}
    \end{subfigure}%
    ~ 
    \begin{subfigure}[t]{0.35\textwidth}
        \centering
        \includegraphics[width=\textwidth]{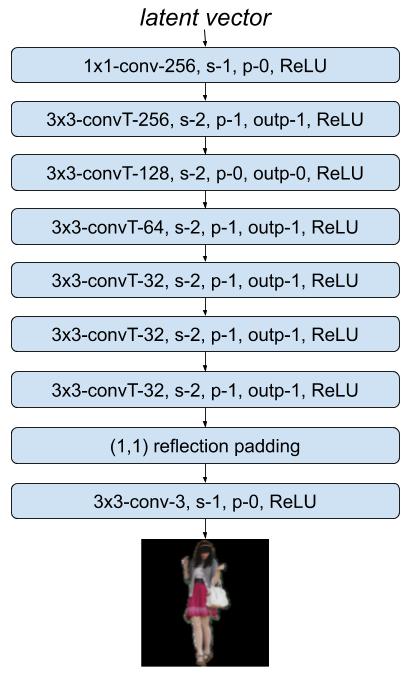}
        \caption{Decoder topology.}
        \label{fig:decoder_topology}
    \end{subfigure}
    \caption{\textbf{The architecture of person appearance encoder and decoder.} We use the following abbreviations: \emph{p}=padding, \emph{s}=stride, \emph{outp}=output padding.}
    \label{fig:appearance_vae}
\end{figure*}

The person appearance encoder introduced in section ``Proposed architecture and loss'' of the main paper is trained as a part of a variational autoencoder, using $64\times64$ crops with humans sourced from the MS-COCO dataset. Its diagram is shown in Figure~\ref{fig:encoder_topology}.
The autoencoder is trained with  Kullback-Leibler (KL) divergence and Cross-Entropy loss. We use Adam \cite{KingmaB14Adam} as an optimizer.
The encoder is implemented as a convolutional network with fully-connected heads and consists of two parts. Its first part is shared by both branches and is fully convolutional. Two fully-connected heads form the second part that outputs mean $\mu$ and variance $\sigma^2$ vectors.  
Given an input image of size $64\times64$ pixels, the encoder produces two $16$-dimensional vectors corresponding to the mean and variance (assumed diagonal) of the approximate latent posterior. We combine them with the reparametrization trick~\cite{Kingma-ICLR-2014} and obtain a latent appearance vector.

The decoder is used only during the training of the variational autoencoder and is not used in our final model. It is implemented as a fully-convolutional network and takes as an input the latent 
appearance vector. Its architecture is given in Figure~\ref{fig:decoder_topology}.

\section{Conditional generator}
\label{sec:conditional_generator}

\begin{figure*}[ht!]
    \centering
    \begin{subfigure}[t]{0.4\textwidth}
        \centering
        \includegraphics[width=0.9\textwidth]{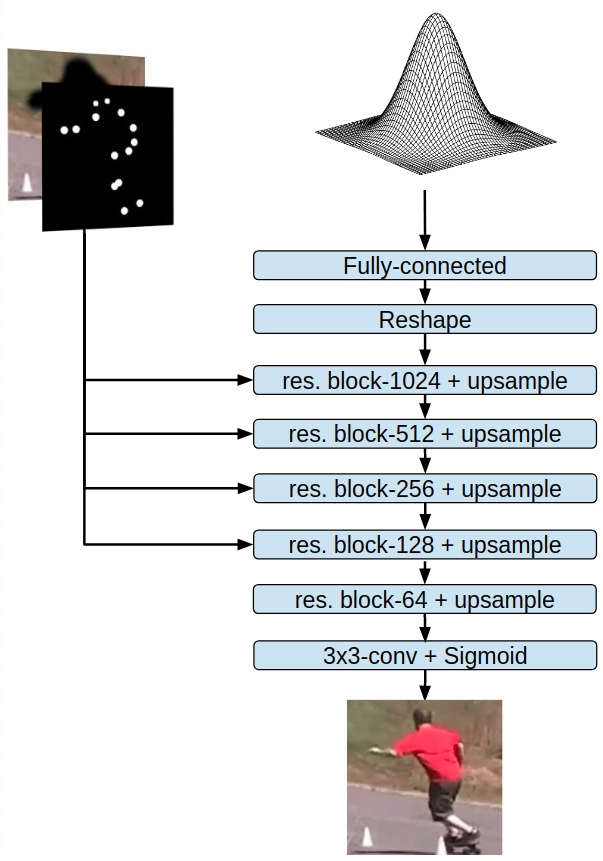} 
        \caption{Overall topology of conditional generator.}
        \label{fig:gen_overall}
    \end{subfigure}%
    \begin{subfigure}[t]{0.5\textwidth}
        \centering
        \includegraphics[width=1.0\textwidth]{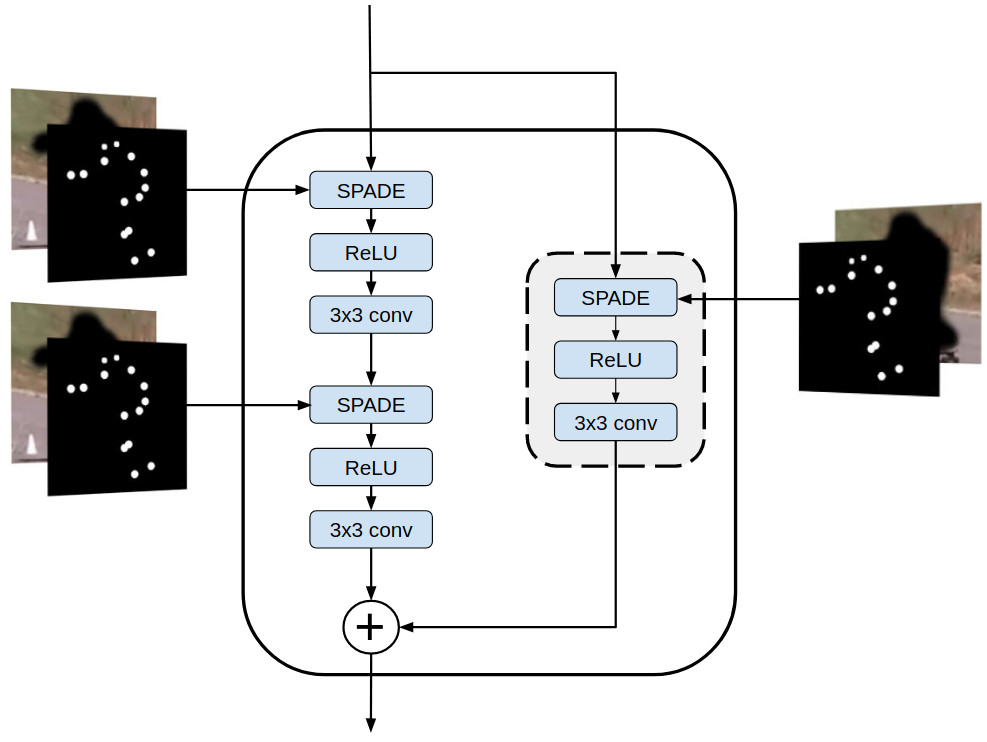} 
        \caption{Close-up on the residual block.}
        \label{fig:res_block}
    \end{subfigure}
    \caption{\textbf{The architecture of the conditional generator.} (a) Overall architecture of the generator. (b) Close-up on the residual block inspired by~\cite{SPADE} showing where the different inputs are fed into the network.}
    \label{fig:generator_topology}
\end{figure*}

The generator takes as input the person appearance latent vector, a background image (with a masked-out pedestrian if there was one) together with keypoints of the target person. The latent appearance vector is the input to the network. 

The background image and target person keypoints, represented as a multi-channel tensor with one channel per keypoint, are concatenated and fed to the network at multiple depths via residual blocks. The details of the topology of the generator, including the locations of the conditioning inputs, are shown in Figure~\ref{fig:generator_topology}.

The generator contains $n$ residual blocks, each followed by bilinear upsampling. This allows us to perform progressive growing of the output with the resulting image size of $s = 16 \cdot 2^{n}$. After these $n$ upsampling blocks, the $N$ resulting feature maps are passed to a $3\times 3$ convolutional layer followed by a sigmoid activation that maps from $N$ to $3$ channels that correspond to RGB color channels. We use $n=4$, which corresponds to the output size of $256 \times 256$ pixels.

\input{include/ablations}

\section{Dataset for DummyNet training and keypoint generation} \label{sec:supp_clust}

In this section, we provide more details about the dataset used for our DummyNet training. We discuss the construction of sets of positive samples of people in various poses, followed by a description of pose clustering and the keypoint generator.

\paragraph{Person image data for DummyNet training.}
We leveraged the YoutubeBB~\cite{RealSMPV17} dataset that contains a large number of videos with people in a large variety of poses, illuminations, resolutions, etc.
We used OpenPose~\cite{Cao2018OpenPose}
to automatically detect people and annotate skeleton keypoints. This way we have annotated all frames from $2,064$ videos from detection folders \emph{person} ($476$ videos) and \emph{skateboard} ($1,588$ videos); specially the latter contains huge variations in poses. We have used $1,911$ videos from $2,064$ annotated ones after filtering out too blurry/dark images. We have also required at least $6$ out of $17$ keypoints visible with at least one hip and one shoulder keypoint visible. Images not satisfying this condition were not used. To train our DummyNet model, we annotated the masks for YoutubeBB automatically by our mask estimation network.

The collected dataset contains $769,176$ frames with automatically annotated keypoints and masks.
The TRN and TST splits for DummyNet training only are done at the videos level and share similar pose distributions. The TRN set contains $612,217$ samples, and the TST set $156,959$ samples. $20\%$ of the TRN set, consisting of samples satisfying minimal height of $190$ px, is used for DummyNet training. The TST set was used for visual control of generated samples' quality.

\paragraph{Pose clustering and keypoint generation.}
To have control over the distribution of poses in the resulting train/test datasets, we group the input images into clusters depicting similar poses. The following two-stage clustering algorithm achieves this.
First, we form clusters with similar visibility of skeleton keypoints. Various human body poses generate different patterns of self-occlusions and render any attempt to measure the distance of two skeletons very difficult. To address this problem, we have a binary \emph{visibility vector} with $17$ entries, each corresponding to one joint for every annotated skeleton.
Clustering on the visibility vectors resulted in $189$ clusters with similar numbers of points and roughly corresponding to different viewpoints. We call this first  stage \emph{viewpoint clustering} and its results \emph{viewpoint clusters}. Every viewpoint cluster has a different visibility vector encoding missing bodyparts (limbs, etc.). Given such training data, the DummyNet is then able to synthesize person images with specific bodyparts occluded.

\begin{figure*}
     \centering
     \begin{subfigure}{0.23\textwidth}
     \centering
    \includegraphics[width=\linewidth]{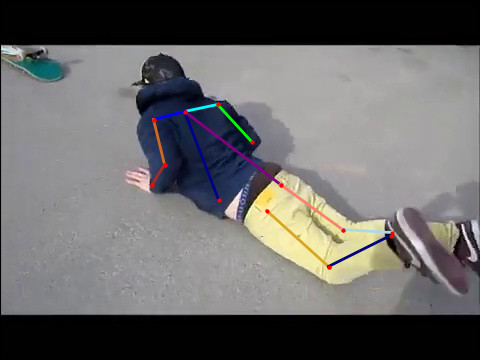} 
     \end{subfigure}
     \begin{subfigure}{0.23\textwidth}
     \centering
    \includegraphics[width=\linewidth]{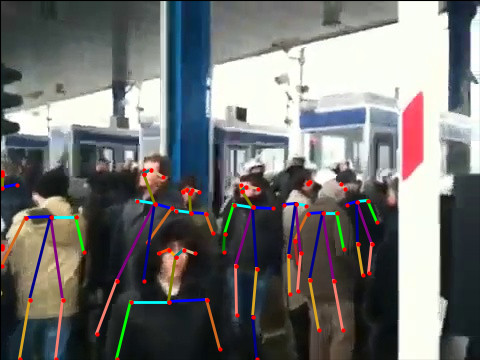}
     \end{subfigure}
     \begin{subfigure}{0.23\textwidth}
     \centering
     \includegraphics[width=\linewidth]{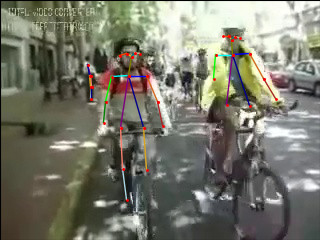}
     \end{subfigure}
     \begin{subfigure}{0.23\textwidth}
     \centering
    \includegraphics[width=\linewidth]{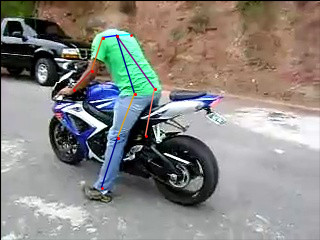}
     \end{subfigure}
     \begin{subfigure}{0.23\textwidth}
     \centering
    \includegraphics[width=\linewidth]{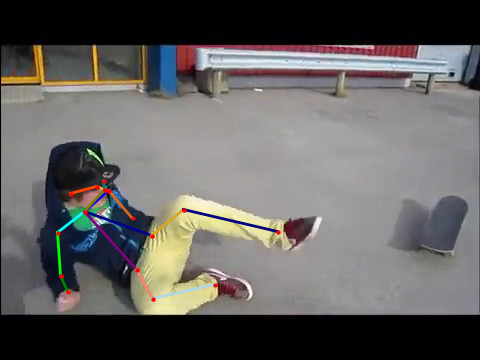} 
     \end{subfigure}
     \begin{subfigure}{0.23\textwidth}
     \centering
    \includegraphics[width=\linewidth]{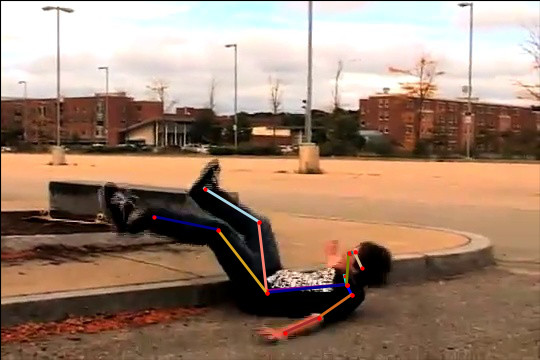} 
     \end{subfigure}
     \begin{subfigure}{0.23\textwidth}
     \centering
    \includegraphics[width=\linewidth]{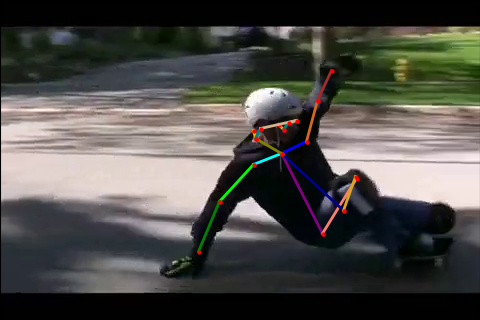} 
     \end{subfigure}
     \begin{subfigure}{0.23\textwidth}
     \centering
    \includegraphics[width=\linewidth]{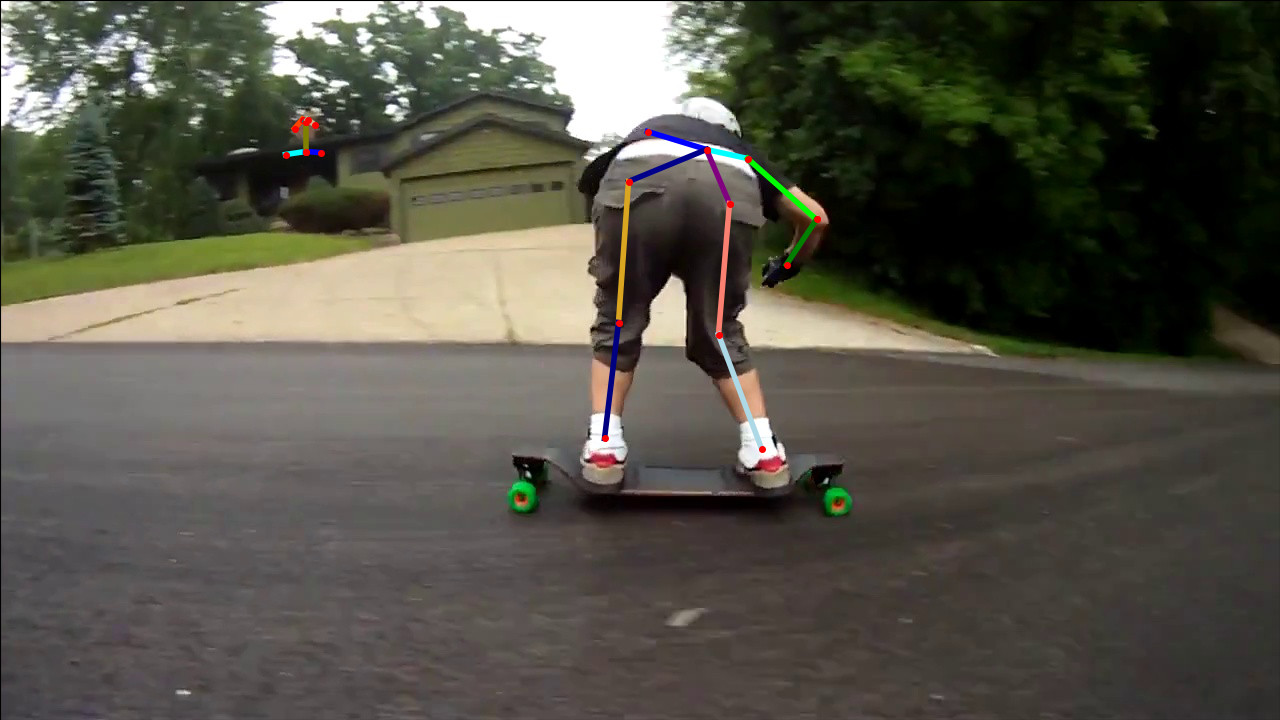} 
     \end{subfigure}
     \caption{\textbf{DummyNet training data}. Samples of the training data with keypoints annotated by OpenPose~\cite{Cao2018OpenPose}.}
     \label{fig:skeletons2}
 \end{figure*}

Second, we cluster poses within each viewpoint cluster. For \emph{pose clustering}, we normalize every skeleton keypoints by subtracting the central torso point (mean of the center of shoulders and the center of hips) and dividing by the torso height (norm of the vector created from the same center points). 
Each skeleton is represented as a $34$-dimensional vector by concatenating $17$ (normalized) $x$- and $y$-coordinates. 
In both clustering stages, we use the Birch clustering method~\cite{Zhang96birch} for its efficiency on large datasets. Examples of DummyNet training images with detected keypoints are depicted in Fig.~\ref{fig:skeletons2}.

\paragraph{Keypoint generator.}

\begin{figure*}[tb]
    \centering
    \begin{subfigure}[t]{0.12\textwidth}
        \centering
        \includegraphics[height=3cm]
        {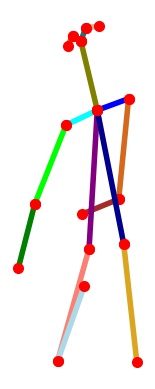} 
    \end{subfigure}%
    \begin{subfigure}[t]{0.12\textwidth}
        \centering
        \includegraphics[height=3cm]
        {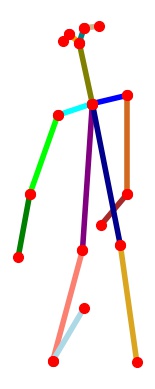} 
    \end{subfigure}%
    \begin{subfigure}[t]{0.12\textwidth}
        \centering
        \includegraphics[height=3cm]
        {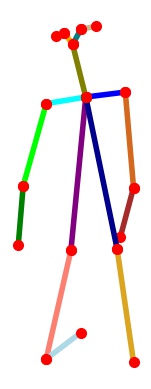} 
    \end{subfigure}%
    \begin{subfigure}[t]{0.12\textwidth}
        \centering
        \includegraphics[height=3cm]
        {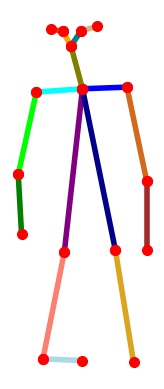} 
    \end{subfigure}%
    \begin{subfigure}[t]{0.12\textwidth}
        \centering
        \includegraphics[height=3cm]
        {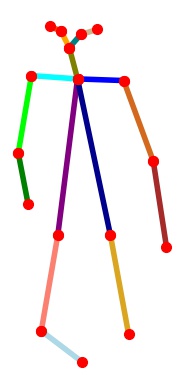} 
    \end{subfigure}%
    \begin{subfigure}[t]{0.12\textwidth}
        \centering
        \includegraphics[height=3cm]
        {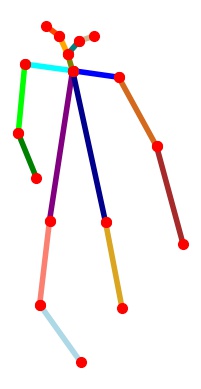} 
    \end{subfigure}%
    \begin{subfigure}[t]{0.12\textwidth}
        \centering
        \includegraphics[height=3cm]
        {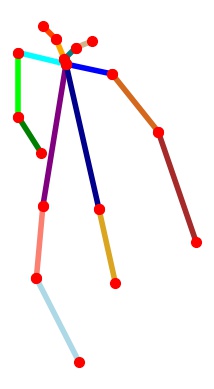} 
    \end{subfigure}%
    \begin{subfigure}[t]{0.12\textwidth}
        \centering
        \includegraphics[height=3cm]
        {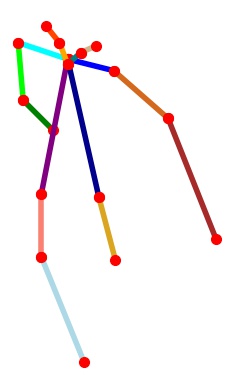} 
    \end{subfigure}
    \begin{subfigure}[t]{0.14\textwidth}
        \centering
        \includegraphics[height=2.6cm]
        {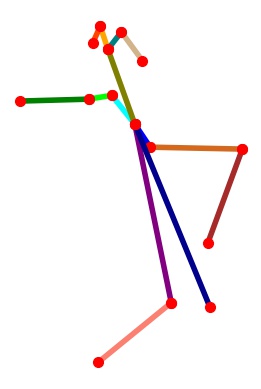} 
    \end{subfigure}%
    \begin{subfigure}[t]{0.14\textwidth}
        \centering
        \includegraphics[height=2.6cm]
        {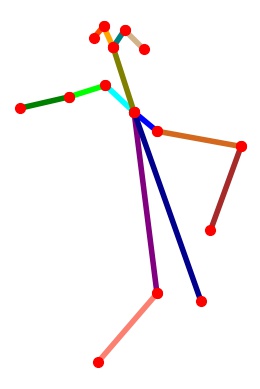} 
    \end{subfigure}%
    \begin{subfigure}[t]{0.14\textwidth}
        \centering
        \includegraphics[height=2.6cm]
        {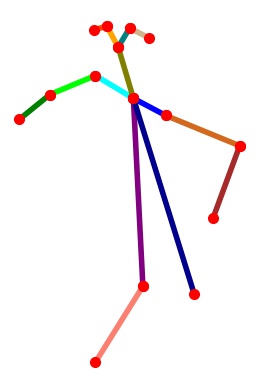} 
    \end{subfigure}%
    \begin{subfigure}[t]{0.14\textwidth}
        \centering
        \includegraphics[height=2.6cm]
        {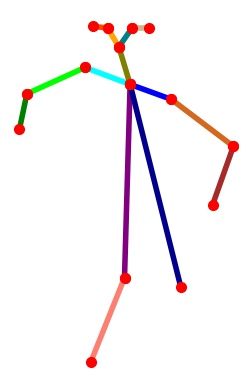} 
    \end{subfigure}%
    \begin{subfigure}[t]{0.14\textwidth}
        \centering
        \includegraphics[height=2.6cm]
        {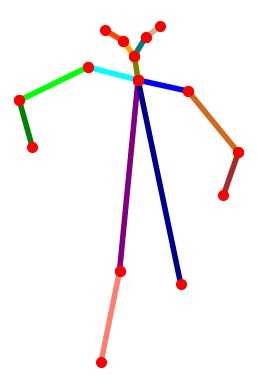} 
    \end{subfigure}%
    \begin{subfigure}[t]{0.14\textwidth}
        \centering
        \includegraphics[height=2.6cm]
        {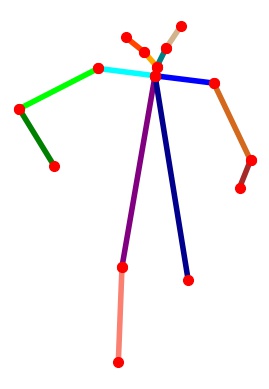} 
    \end{subfigure}%
    \begin{subfigure}[t]{0.14\textwidth}
        \centering
        \includegraphics[height=2.6cm]
        {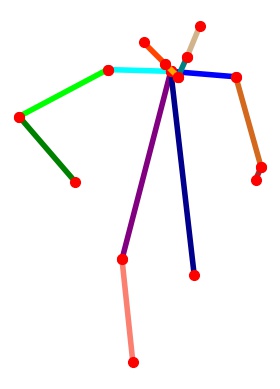} 
    \end{subfigure}
    \caption{\textbf{Generated skeletons with some body parts missing (specified by viewpoint cluster)}. From left to right, the top three principal components (out of 20) are smoothly changing from min to max as parameters controlling the pose. Each row represents one pose cluster.}
    \label{fig:pca}
\end{figure*}

We need to sample various person poses (skeleton keypoints) as the input to the DummyNet illustrated in Figure~\ref{fig:supp_scheme_infer}. This is achieved by learning a simple statistical model on the subset of the extracted person keypoints. 
In detail, for \emph{every pose cluster} which has more than $20$ samples we compute Principal Component Analysis (PCA)~\cite{Jolliffe:1986} decomposition on normalized skeletons ($\mathcal{N}\left( \mathbb{0}, \mathbb{1}\right)$ withing the cluster). The statistical skeleton model then uses the learned cluster-specific PCA matrices and minimal and maximal bounded intervals of the $20$ principal components for sampling new skeletons. At runtime for the selected cluster, we take randomly sampled coefficient values in the corresponding intervals and project them back to the $34$-dimensional space of normalized joints coordinate vectors (See examples in Fig.\,\ref{fig:pca}). Every such vector is then scaled up to fit the $256\times256$ background image at the DummyNet input. This skeleton is used to generate the foreground mask via Mask Estimator. This statistical model allows us to sample person poses that are not available in the training set.

\vspace{-2ex}

\section{Extending Cityscapes scenes}
\label{sec:cs_augmentation}
We use our DummyNet to extend the training set of the Cityscapes dataset. First, to choose the position of the inserted person, we allow it to be placed only either on the ground, road, or sidewalk. Then, we try to place it near an already-present person. We consider only persons that are at least 50 pixels high. If we are not successful, we choose the position randomly. To estimate the height of the person, we fit a linear function $h=a\cdot y + b$ to the training data, where $h$ is person height, and $y$ is a position of the bottom of the person bounding box measured from the bottom of the image. We require that the new person stands on the ground and does not occlude other persons already present in the image that is higher than 50px. This setup is used in the CSP detector experiment (Section~C in the paper). For examples, please see Figure~\ref{fig:cs_fine_hard2}.

\begin{figure*}[tb]
\centering
\begin{subfigure}{1\linewidth}
\centering
\includegraphics[width=0.49\linewidth]{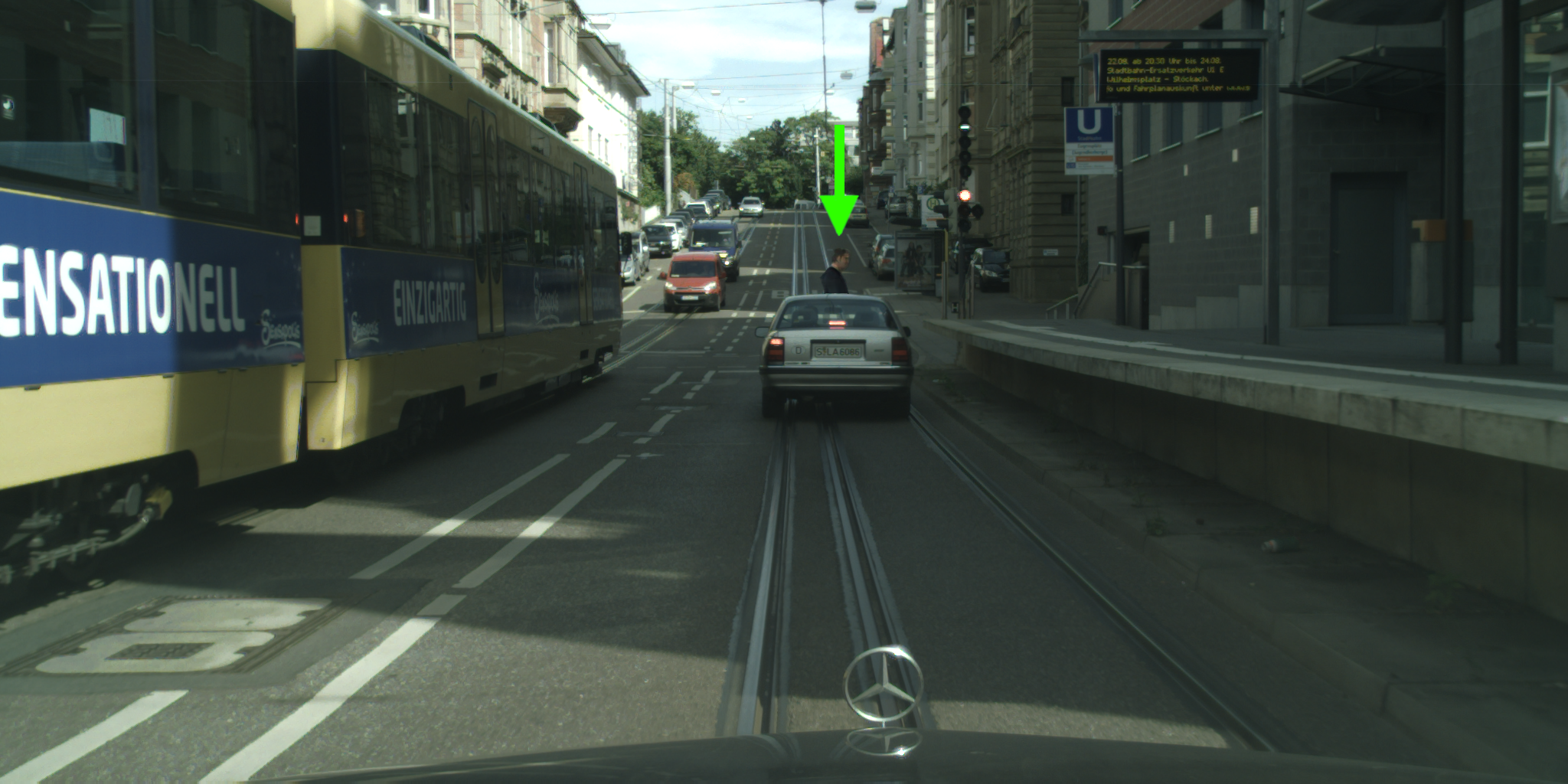}%
\includegraphics[width=0.49\linewidth]{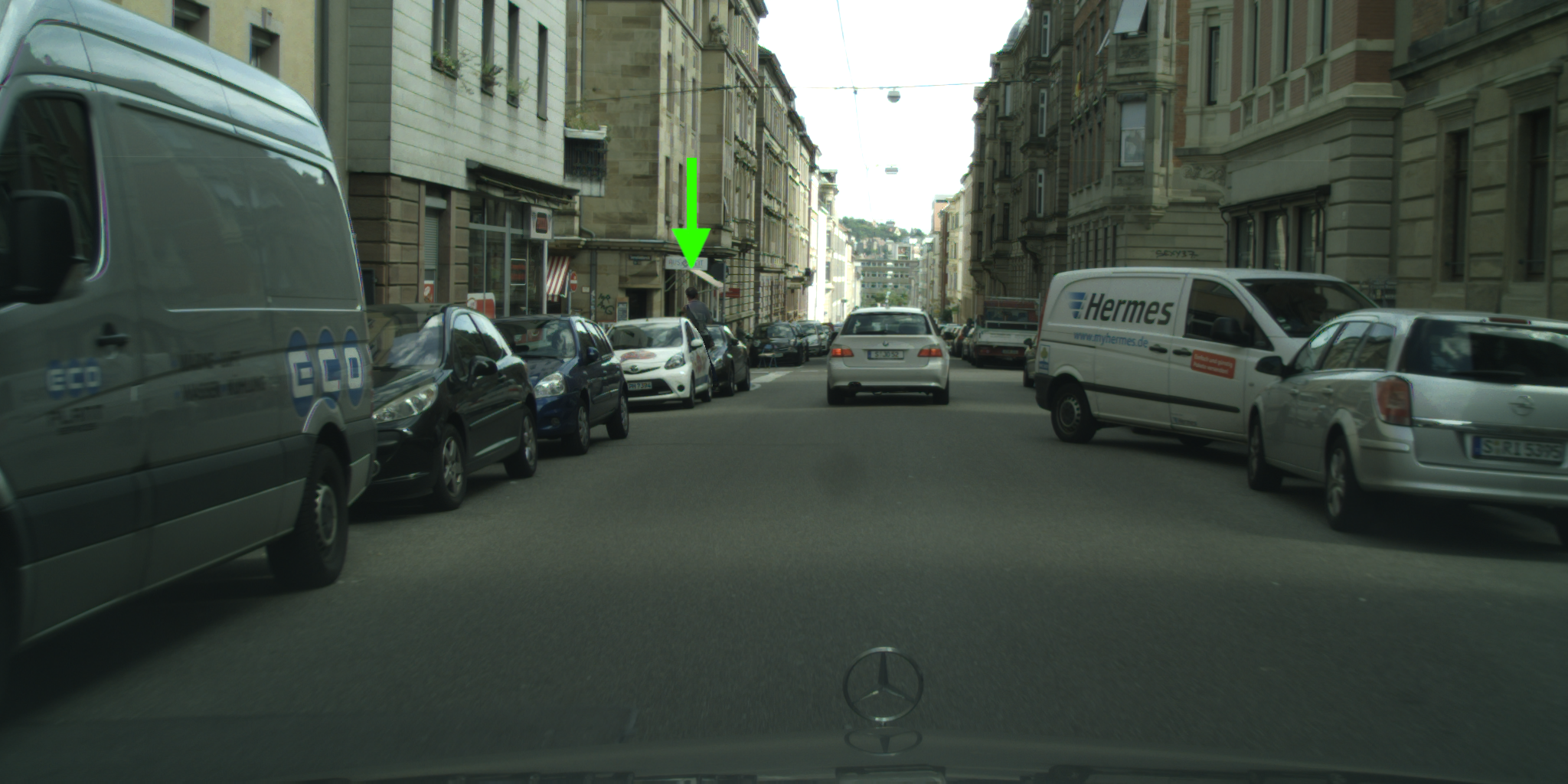}
\end{subfigure}
\begin{subfigure}{1\linewidth}
\centering
\includegraphics[width=0.49\linewidth]{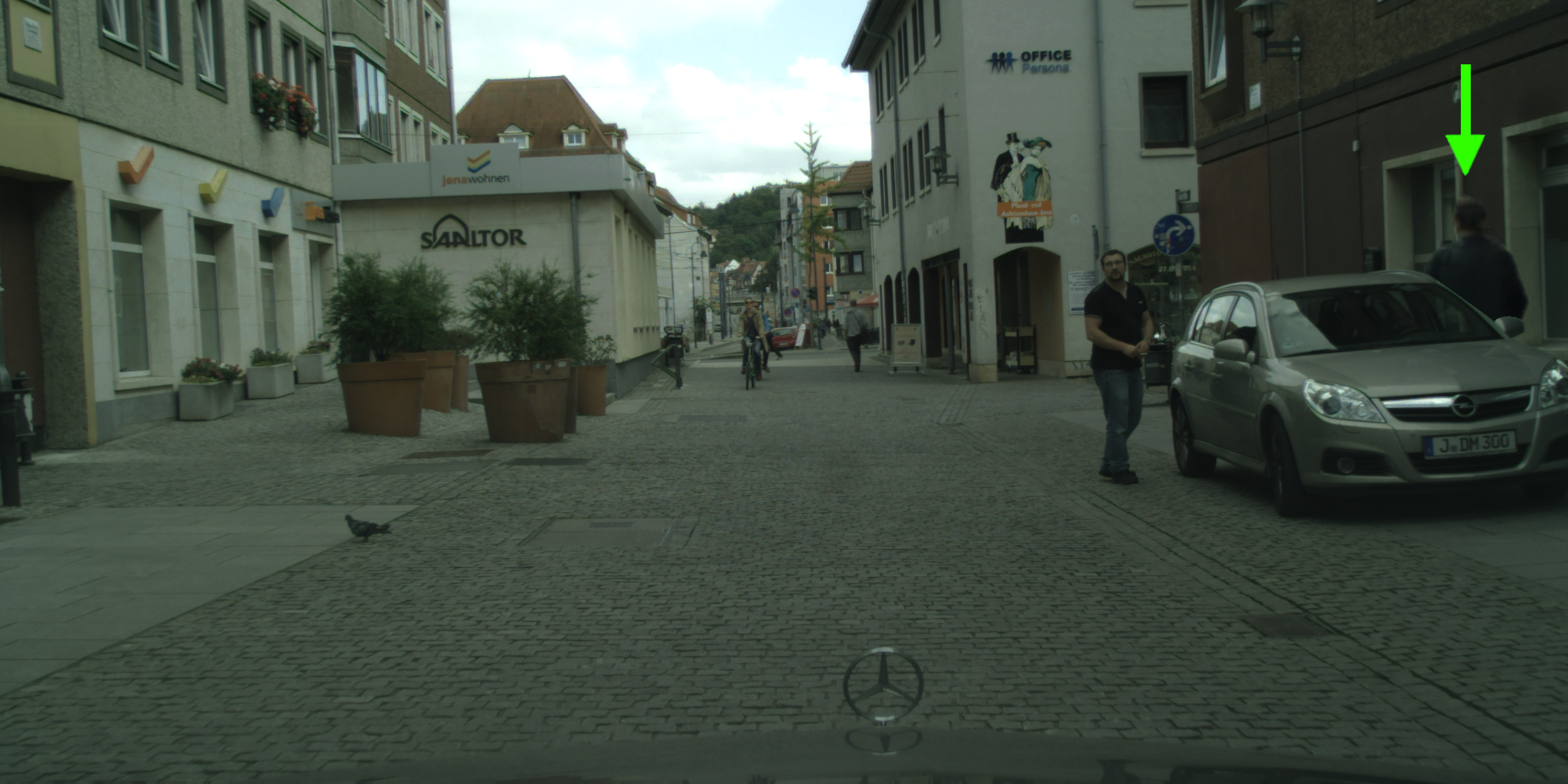}%
\includegraphics[width=0.49\linewidth]{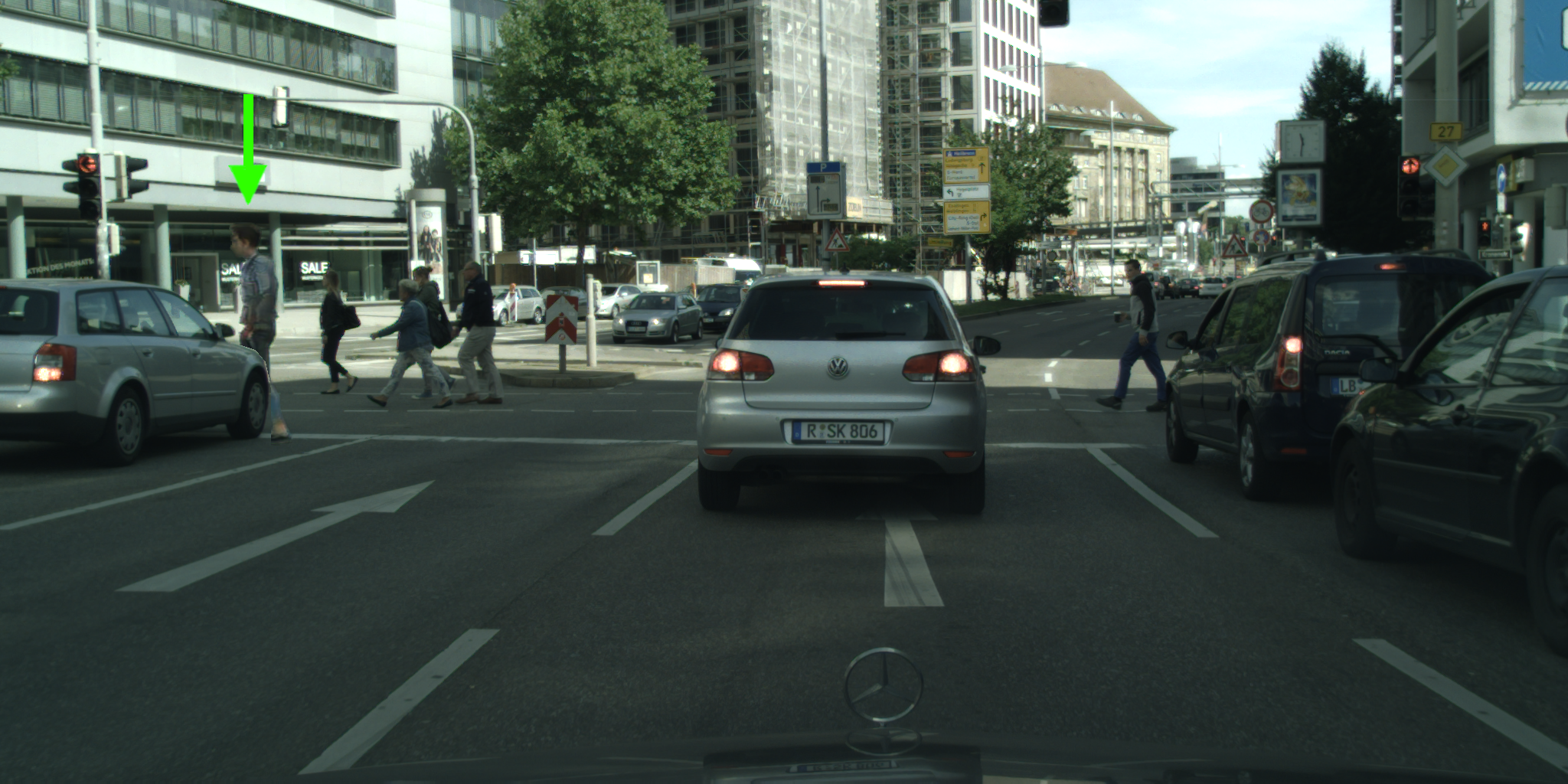}
\end{subfigure}
\begin{subfigure}{1\linewidth}
\centering
\includegraphics[width=0.49\linewidth]{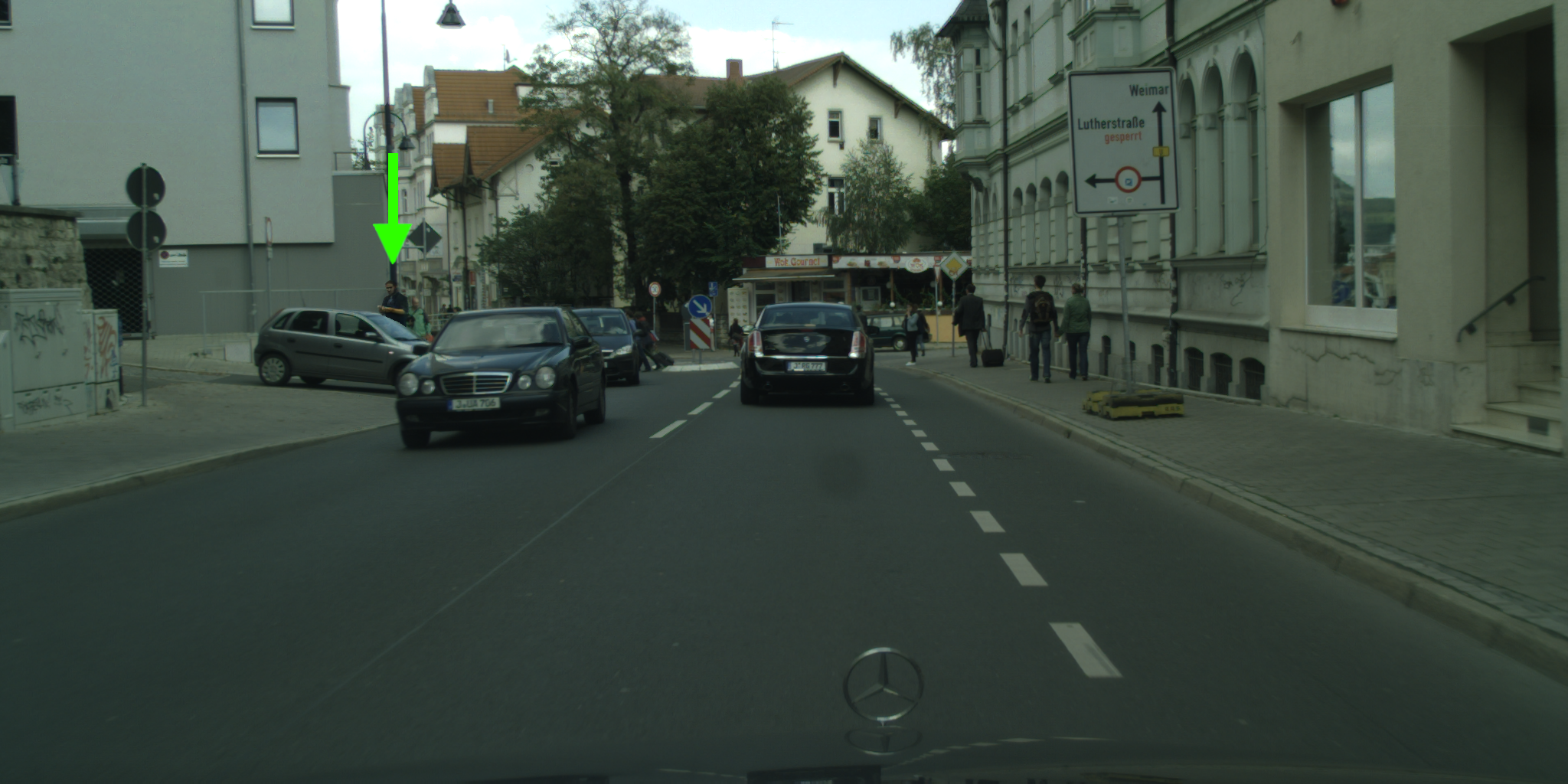}%
\includegraphics[width=0.49\linewidth]{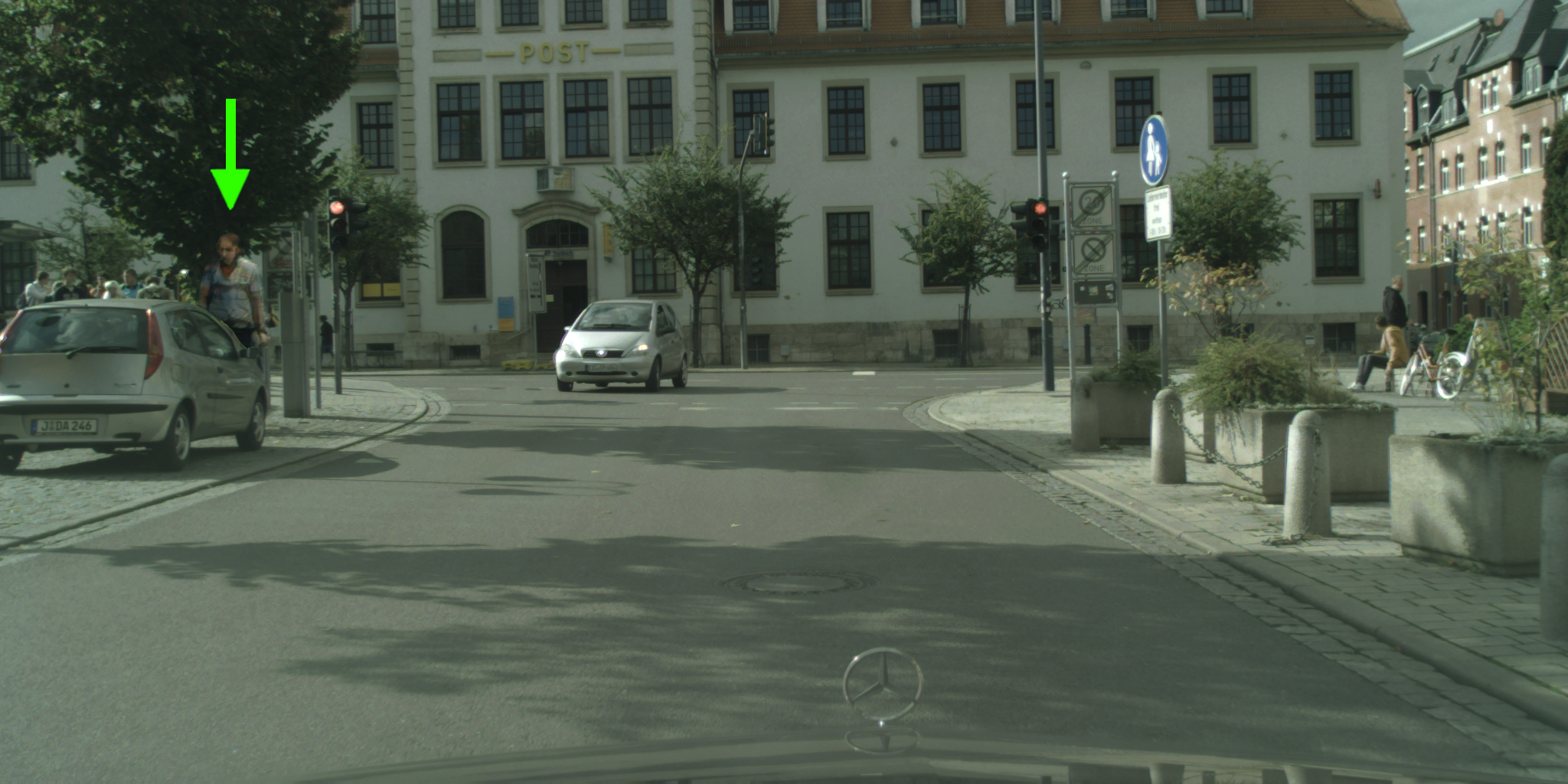}
\end{subfigure}
\begin{subfigure}{1.0\linewidth}
\centering
\includegraphics[width=0.49\linewidth]{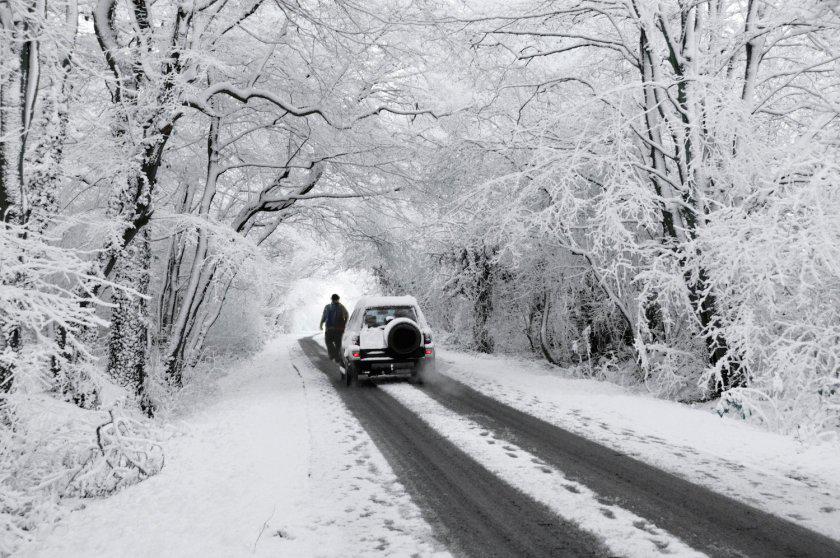}%
\includegraphics[width=0.49\linewidth]{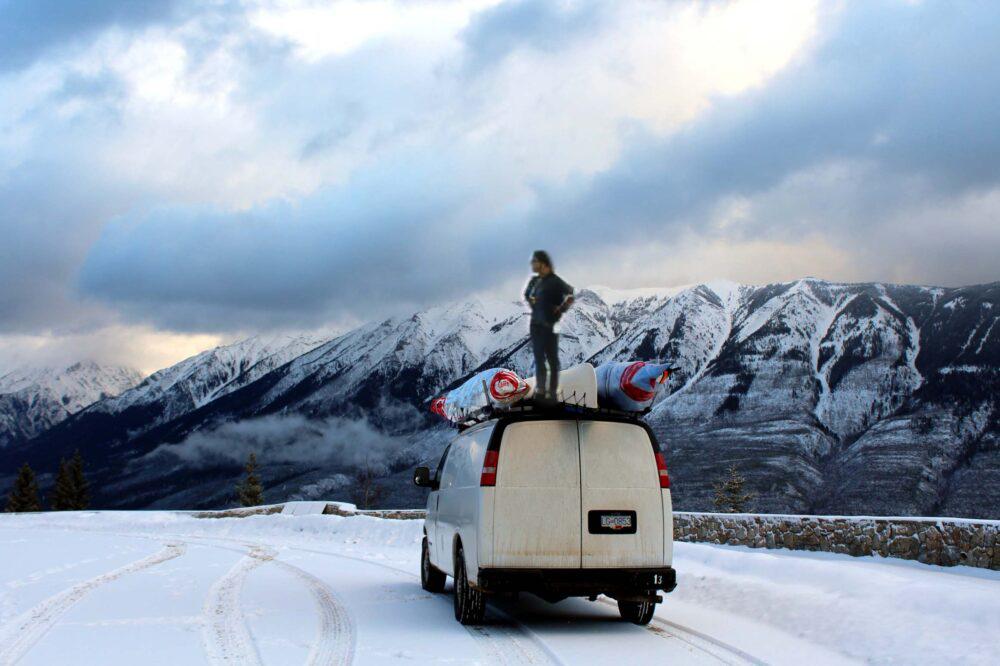}
\end{subfigure}
\caption{{\bf Occluded samples in Cityscapes (rows 1-3) and winter scenes (last row).} 
One synthetic person was added into each image with the exception of the lower right \emph{accident ahead} scene, where three synthetic people were added. These results demonstrate the appearance variability our approach can deal with. These images were produced for this supplementary material to demonstrate the possibilities of our approach and were not used in training the person detectors reported in the main paper.}
\label{fig:occluded_and_winter}
\end{figure*}

Our approach can also generate people occluded by other objects in the scene (although this is already partially covered by our ability to generate person images with occluded bodyparts). We show examples of such synthesized scenes in Figure~\ref{fig:occluded_and_winter}. These images were not yet used for training the person detectors. This approach currently requires semantic segmentation of individual object instances in the scene.

\ifaaai
    \newcommand\w{0.973}
\else
    \newcommand\w{1.0}
\fi

\begin{figure*}[tb]
    \centering
    \begin{subfigure}{\w\linewidth}
    \centering
    \includegraphics[width=0.5\linewidth]{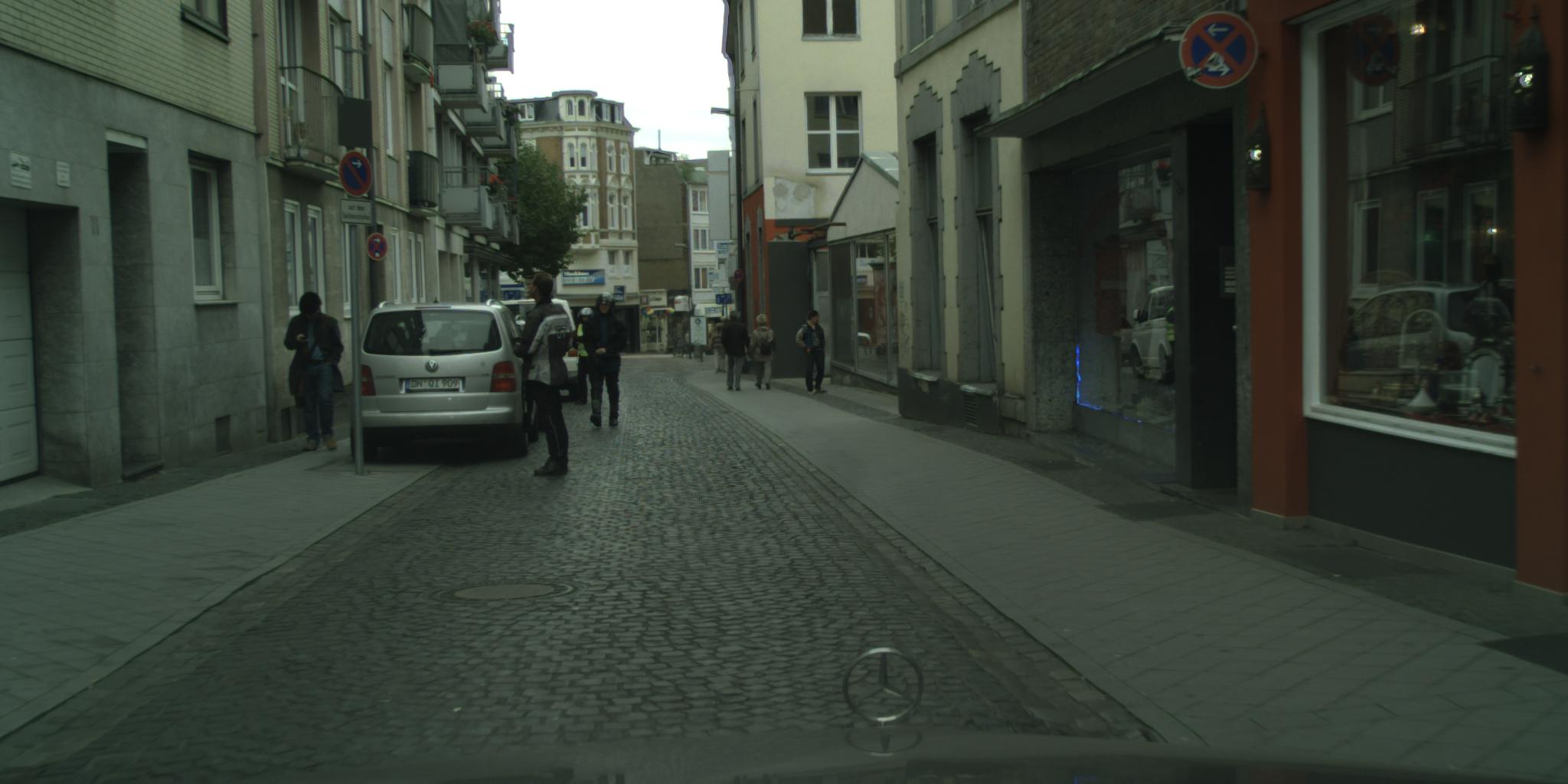}%
    \includegraphics[width=0.5\linewidth]{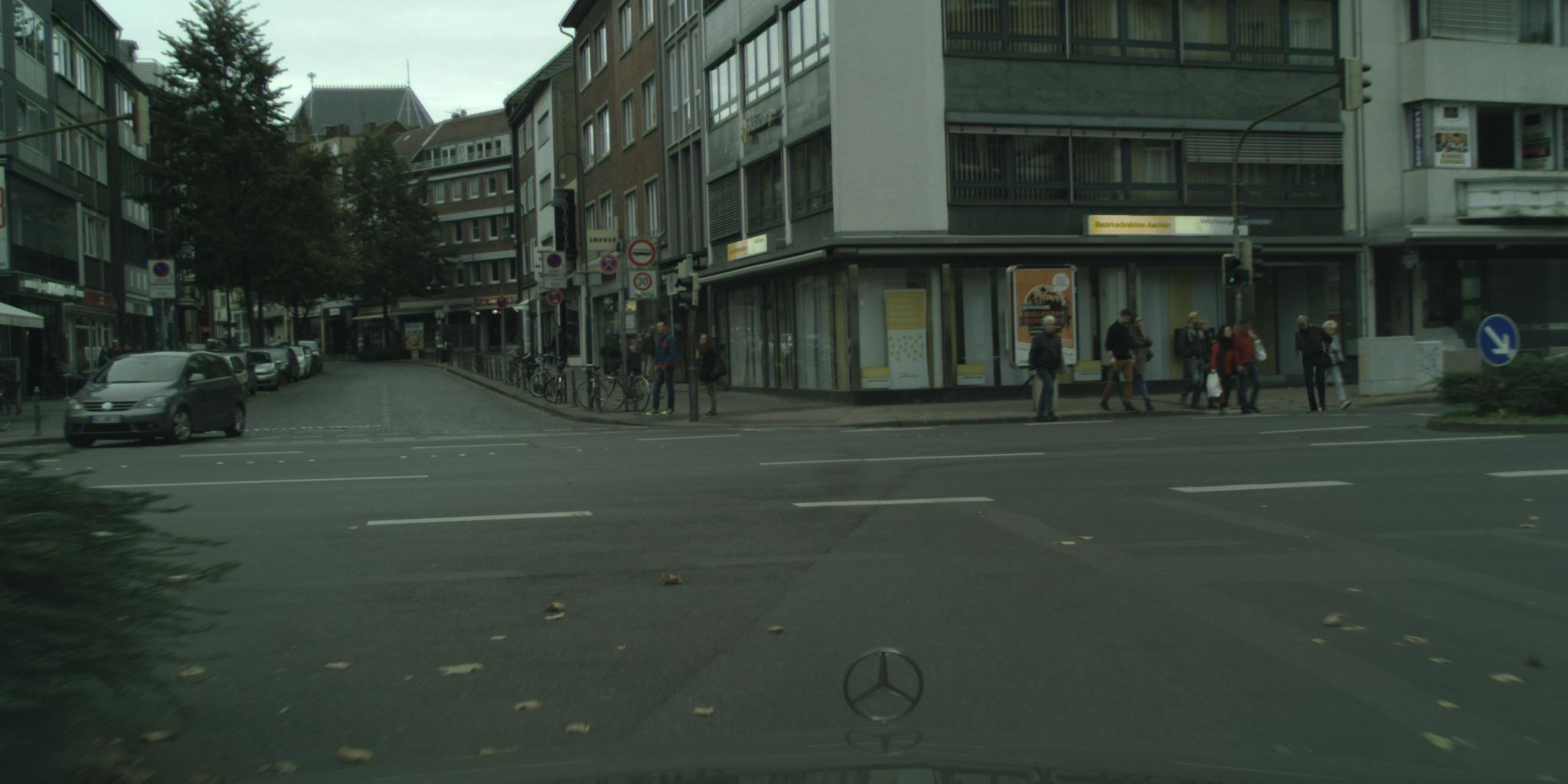}
    \end{subfigure}
    \centering
    \begin{subfigure}{\w\linewidth}
    \centering
    \includegraphics[width=0.5\linewidth]{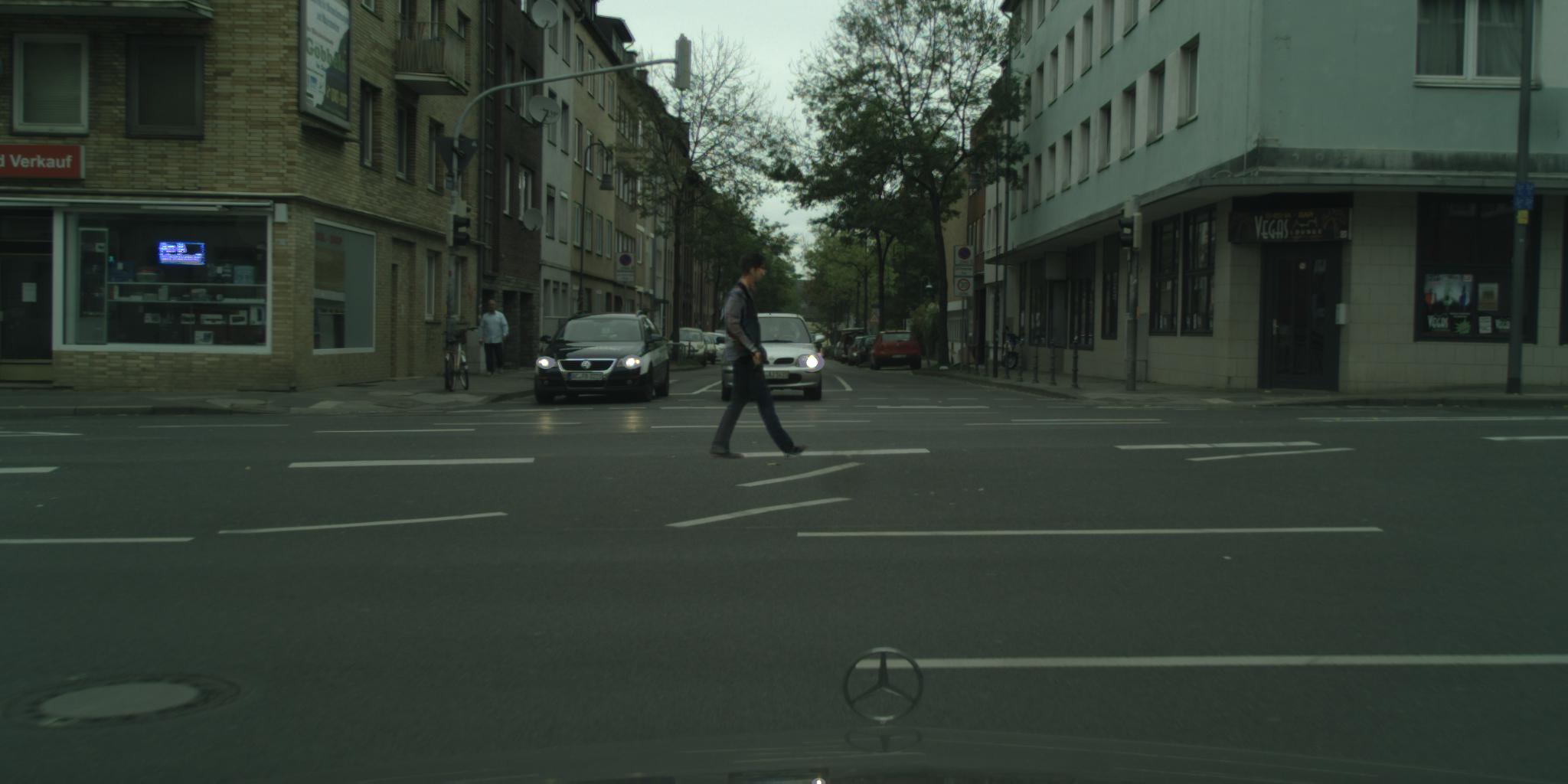}%
    \includegraphics[width=0.5\linewidth]{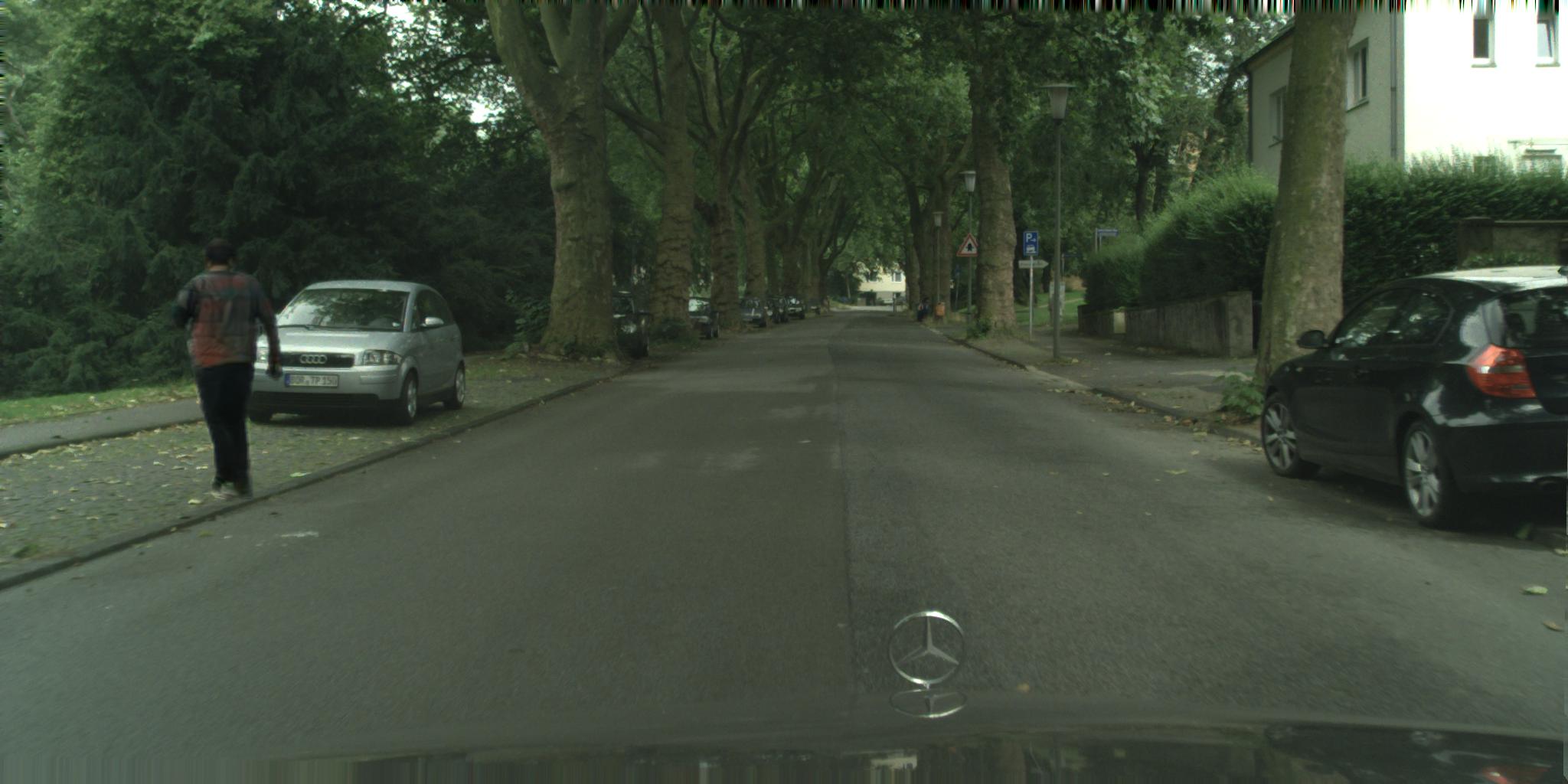}
    \end{subfigure}
    \centering
    \begin{subfigure}{\w\linewidth}
    \centering
    \includegraphics[width=0.5\linewidth]{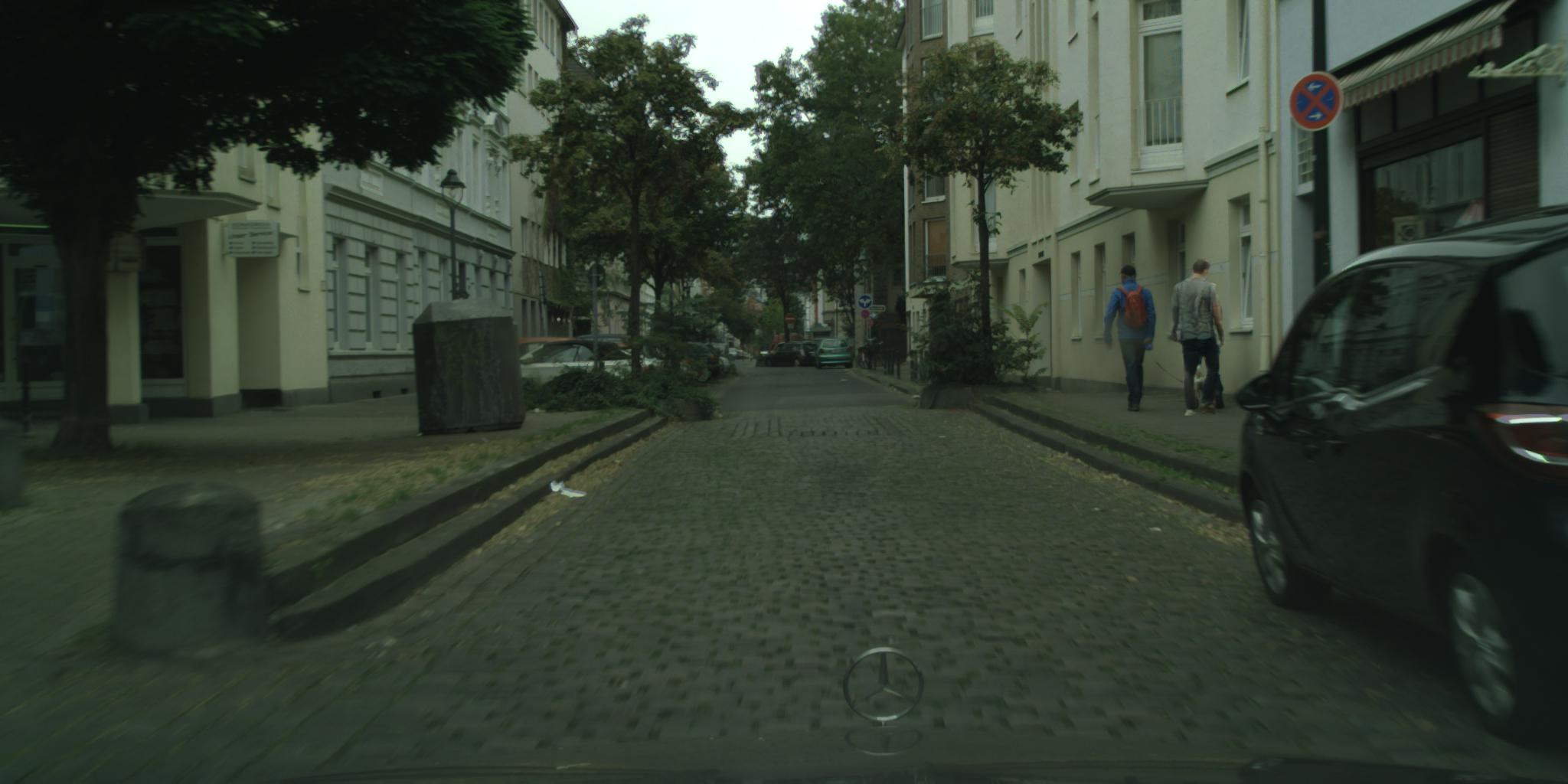}%
    \includegraphics[width=0.5\linewidth]{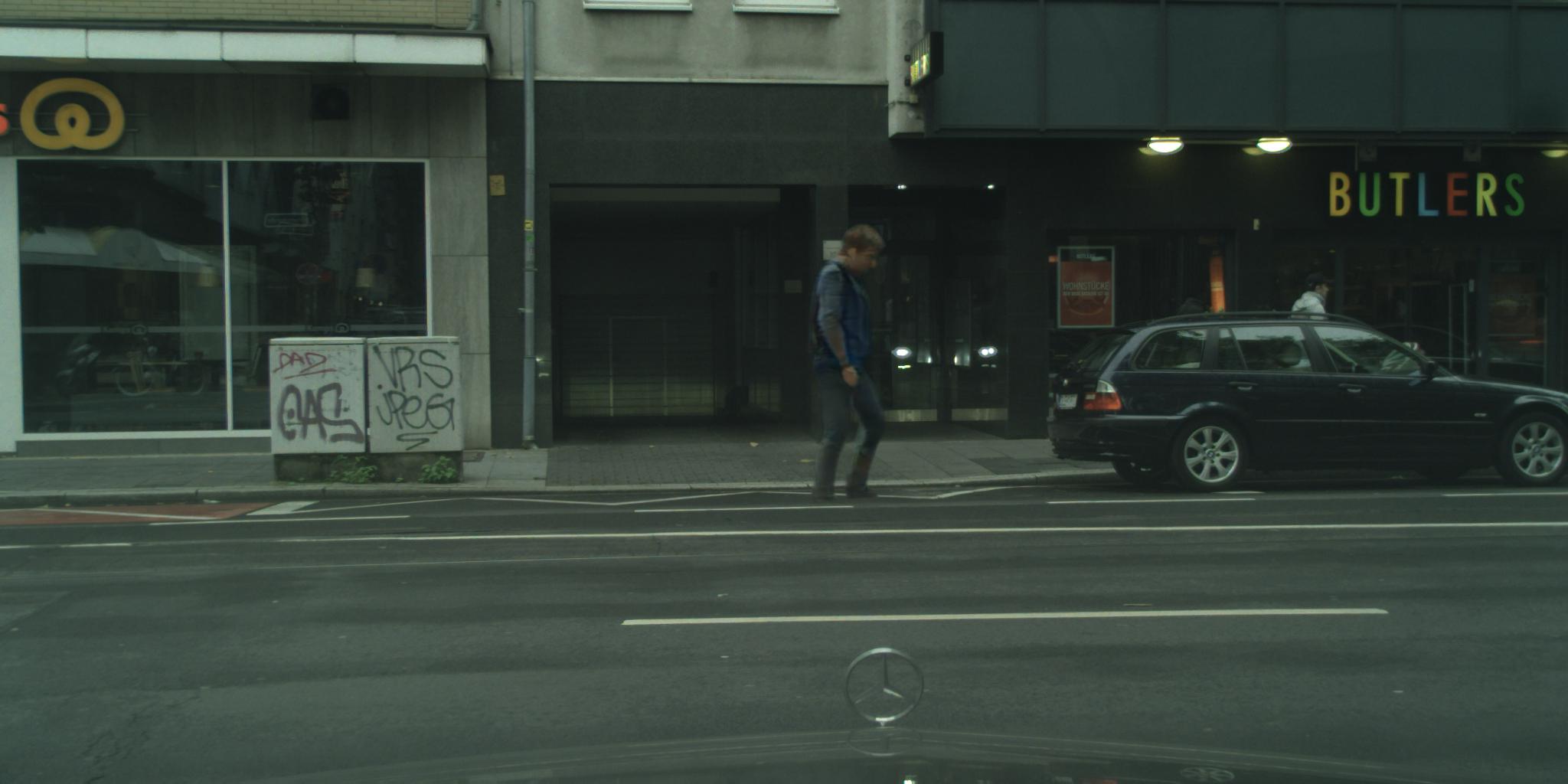}
    \end{subfigure}
    \centering
    \begin{subfigure}{\w\linewidth}
    \centering
    \includegraphics[width=0.5\linewidth]{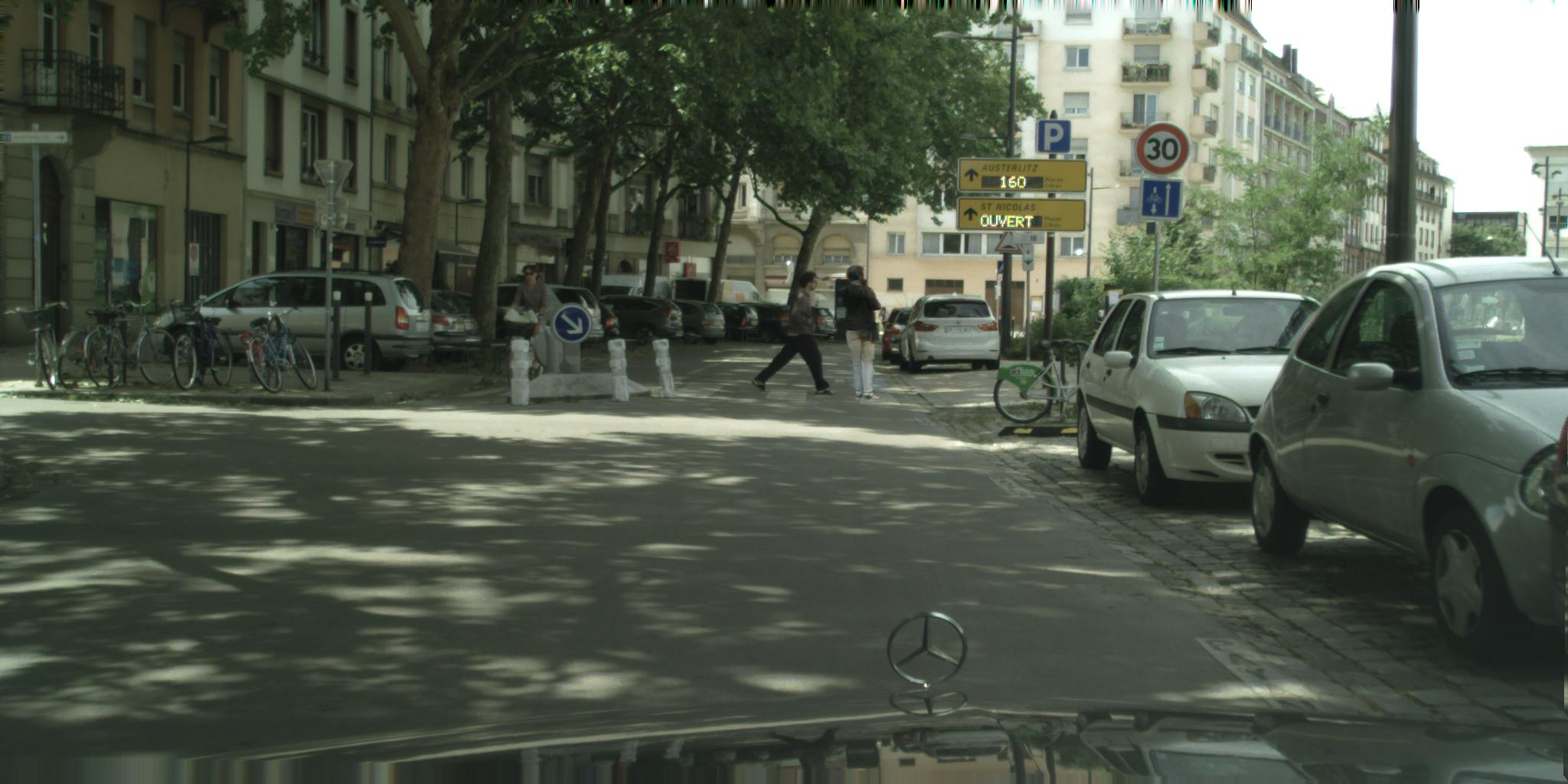}%
    \includegraphics[width=0.5\linewidth]{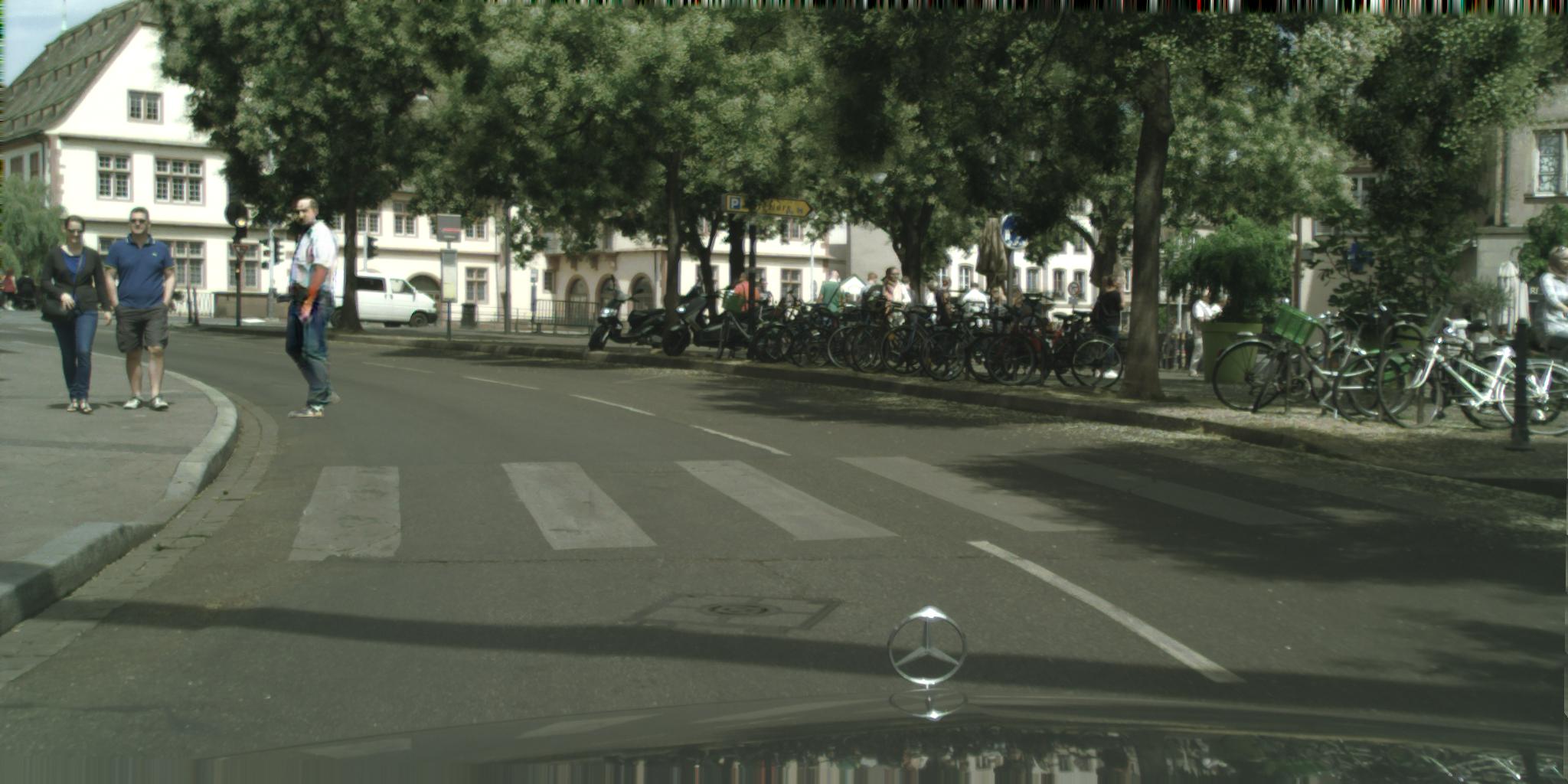}
    \end{subfigure}
    \centering
    \begin{subfigure}{\w\linewidth}
    \centering
    \includegraphics[width=0.5\linewidth]{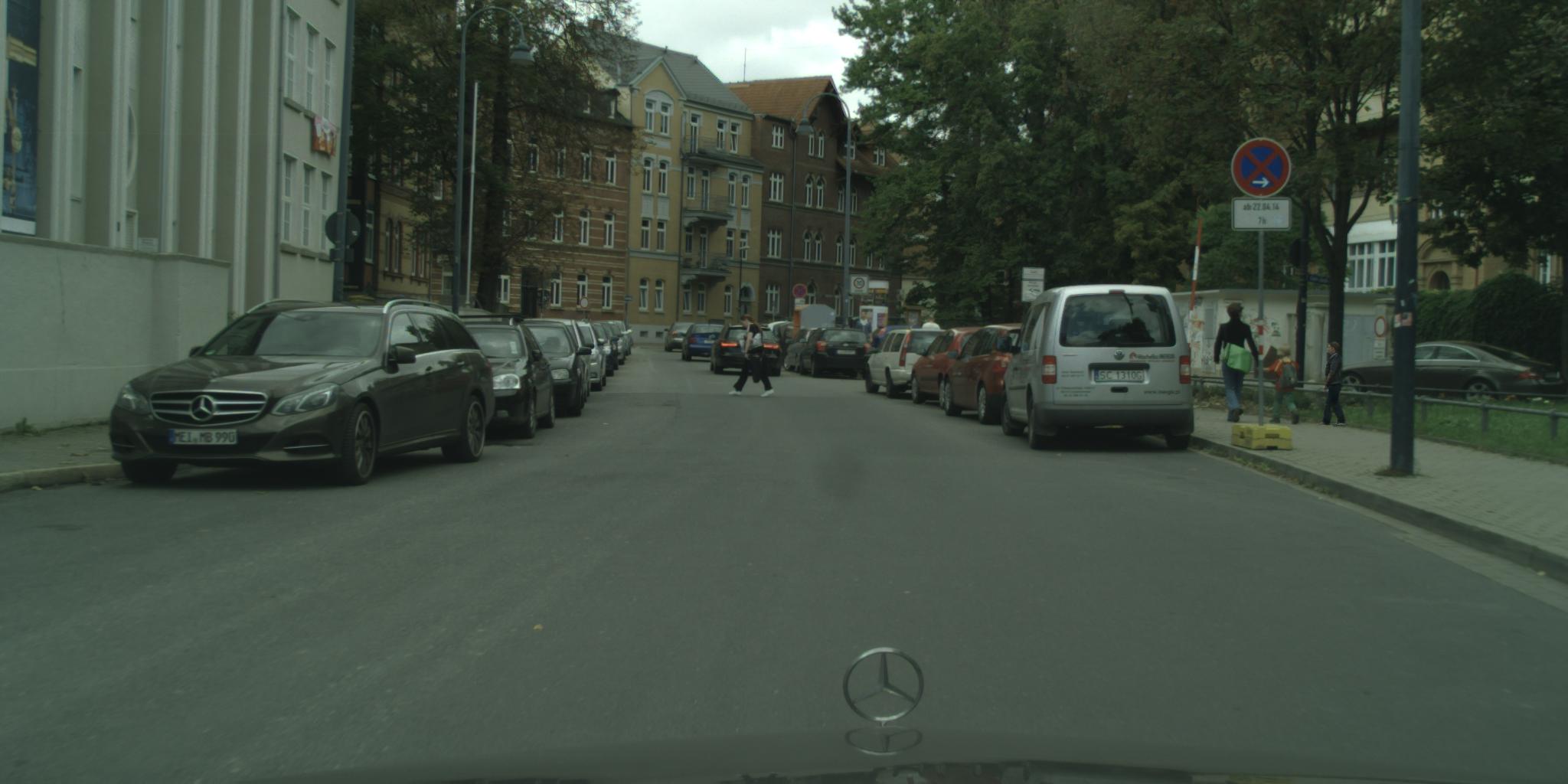}%
    \includegraphics[width=0.5\linewidth]{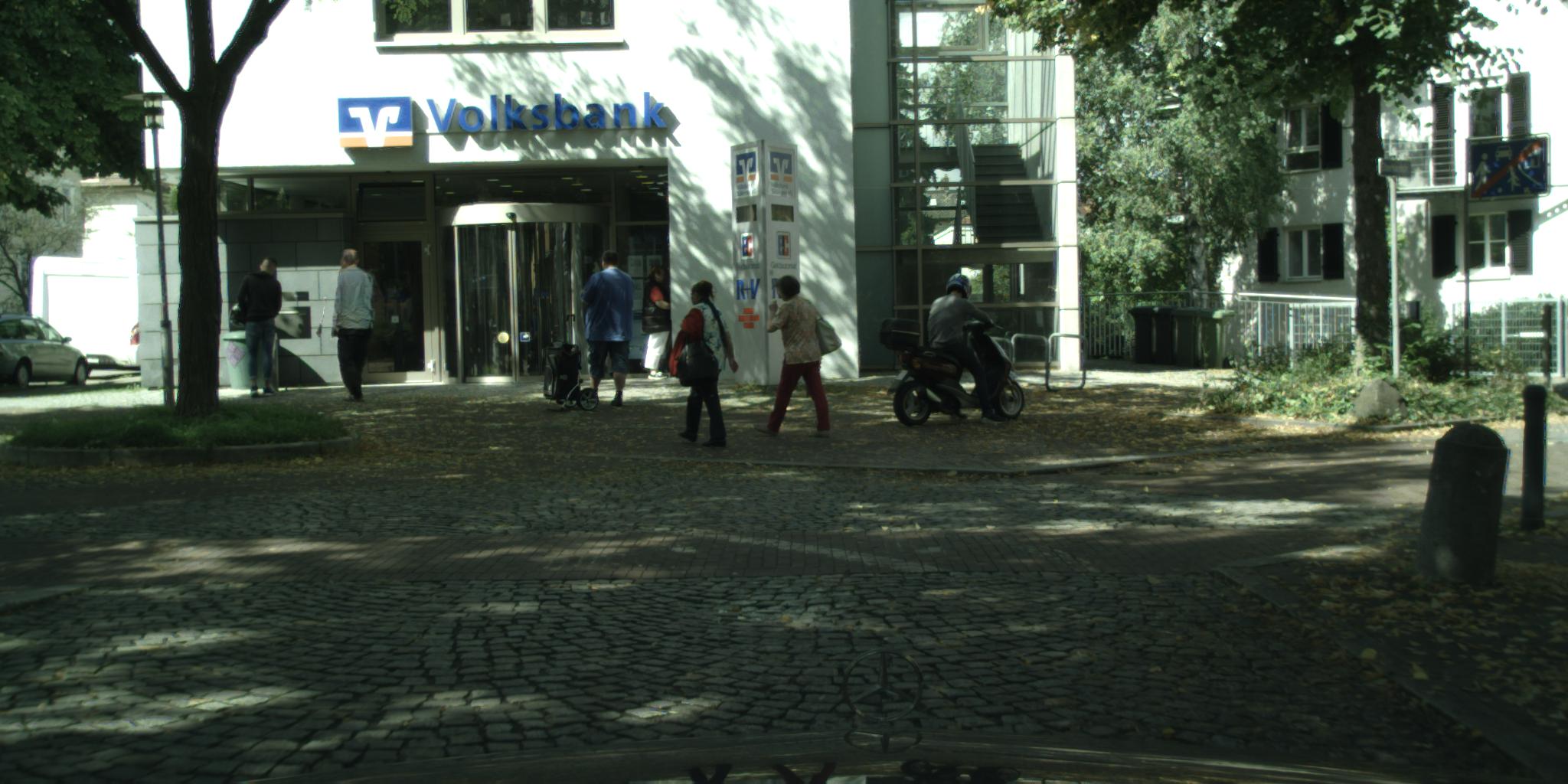}
    \end{subfigure}

    \caption{\small \textbf{Examples of person images generated into Cityscapes scenes}. One synthetic person was added into each scene. 
    See Section~\ref{sec:cs_augmentation} for details of the placement procedure.
    }
    \label{fig:cs_fine_hard2}
\end{figure*}

\section{Training details and additional results}
\label{sec:exp_details}

In this section, we provide more training details and additional results to the experiments from the main paper.

First, in the section~\ref{sec:nightowls}, we provide further information about the experiment B that contains night scenes with annotated people.
Then, in section~\ref{sec:expC_details}, you can find more evaluation results of experiment C. Finally, in section~\ref{sec:expD_details}, you can find training details of experiment~D as well as a complete set of results in all setups available.

\subsection{Person classification at night time}
\label{sec:nightowls}

In this section, we provide further information about the experiment with the NightOwls dataset~\cite{Neumann2018NightOwls} that contains night scenes with annotated people.

For this experiment, we used the annotated training and validation split provided in the dataset.
From the provided splits, we extracted images of pedestrians that were annotated as not occluded and were at least 50px high and rescaled those images to $128\!\times\!128$ px.
We then further divide images from the training split into $12,000$ images for training and $1000$ for validation. All of the $4,886$ original validation images from the NightOwls dataset were kept as a test set.
To get the person-free samples, we sampled the images via the sliding window method in various scales and made sure that they do not contain any annotated person bounding box. In our experiments, we used a fixed set of $50$k negative samples for training, $10$k for validation, and $4,886$ for testing.

We train our classifier using SGD with an initial learning rate of $0.05$ for $1000$ epochs to be sure to reach convergence.
We track the validation metrics and then test the network that performs the best on the validation set. We use binary cross-entropy loss for training. We train $3$ baseline classifiers using $100$, $1$k, and $12$k real positive training samples. 

\begin{figure*}[tb]
\centering
\begin{subfigure}{\w\linewidth}
\centering
\includegraphics[width=0.49\linewidth]{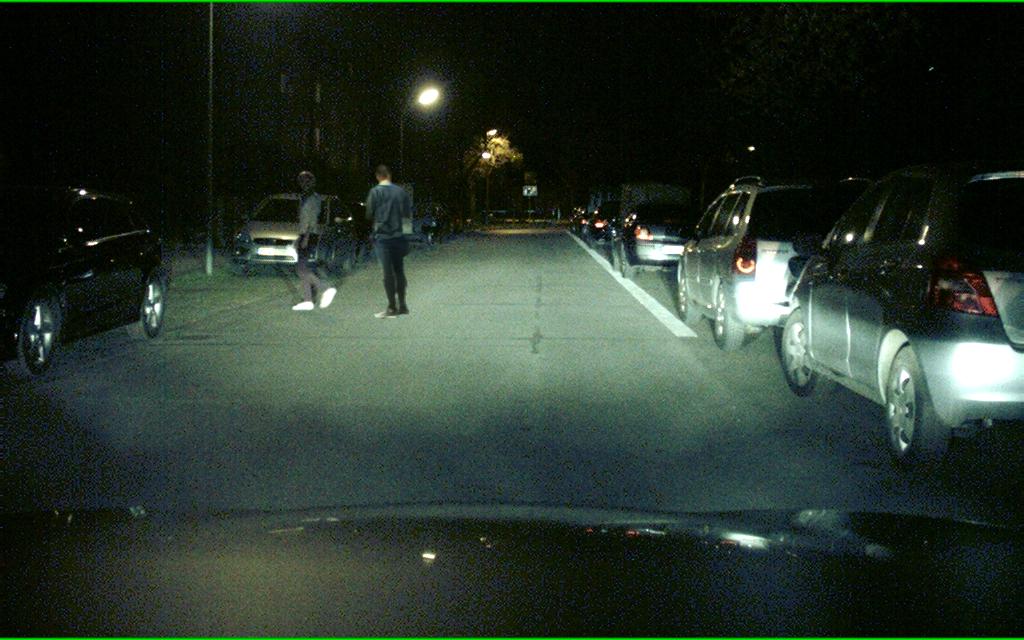}%
\includegraphics[width=0.49\linewidth]{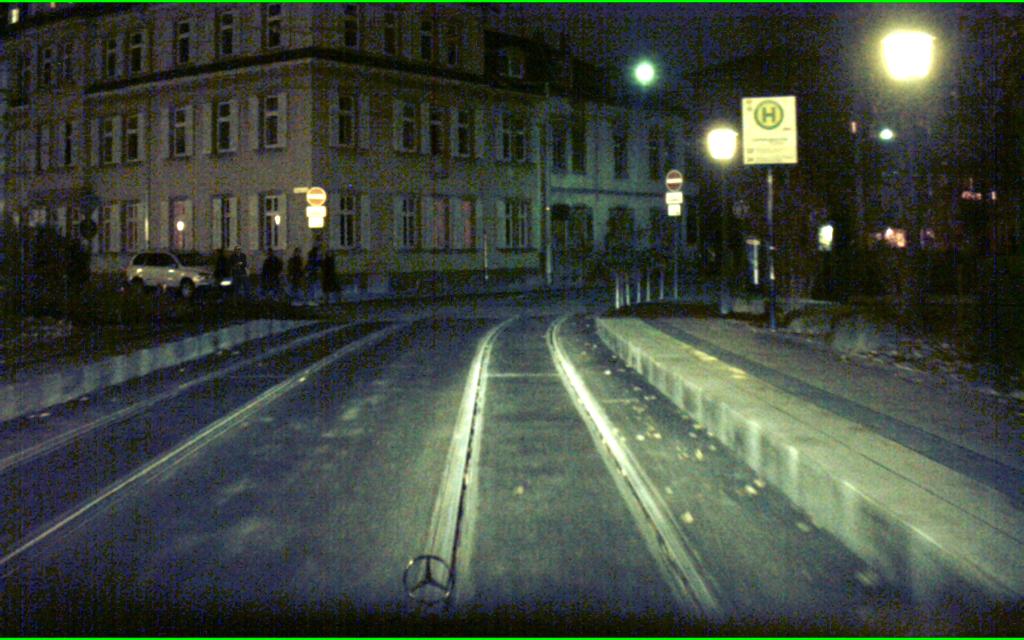}
\end{subfigure}
\begin{subfigure}{\w\linewidth}
\centering
\includegraphics[width=0.49\linewidth]{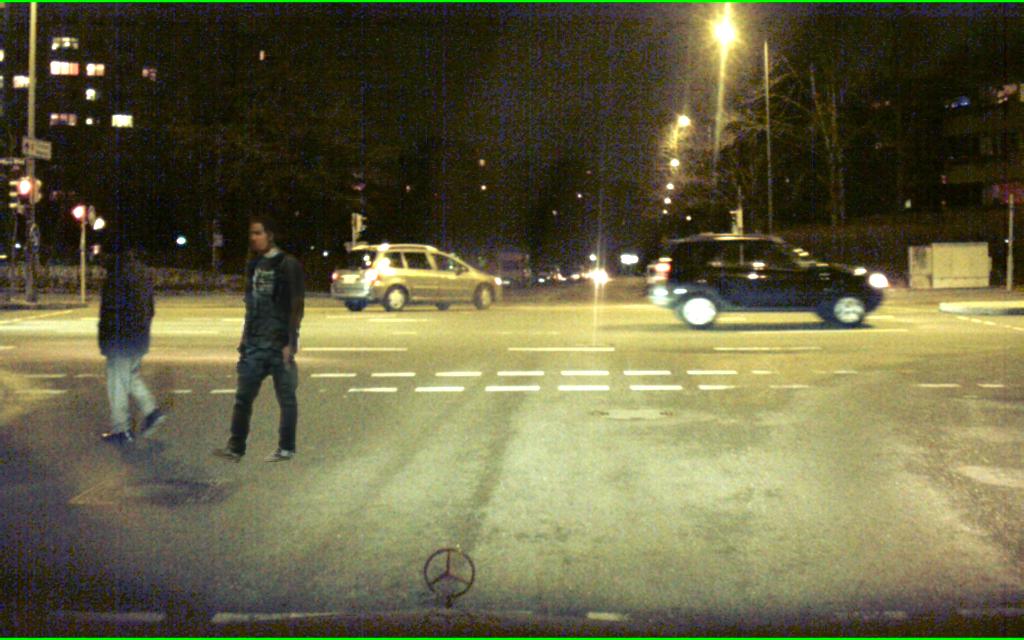}%
\includegraphics[width=0.49\linewidth]{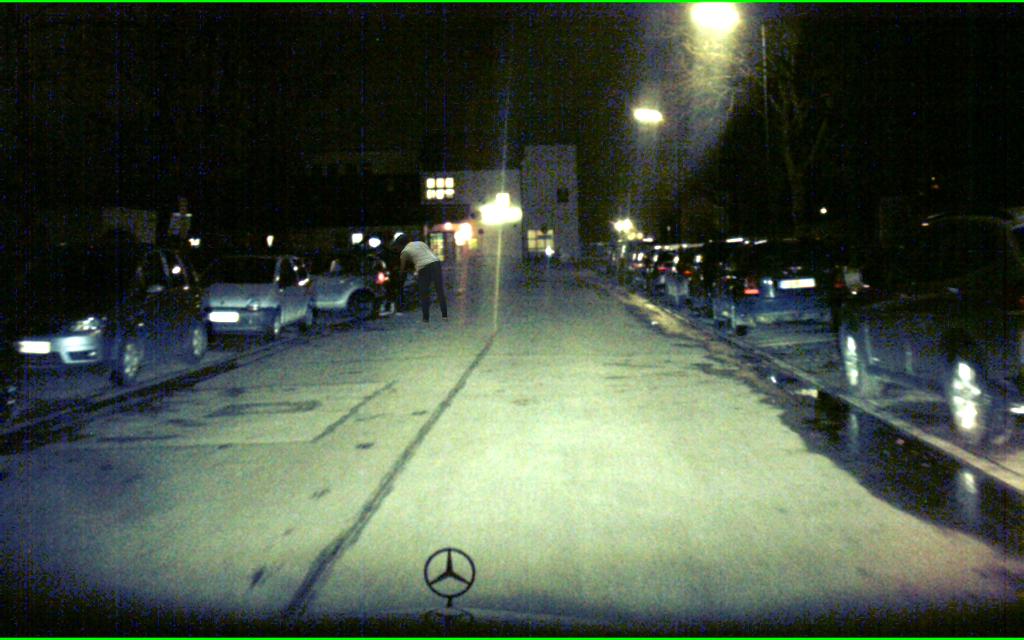}
\end{subfigure}
\begin{subfigure}{\w\linewidth}
\centering
\includegraphics[width=0.49\linewidth]{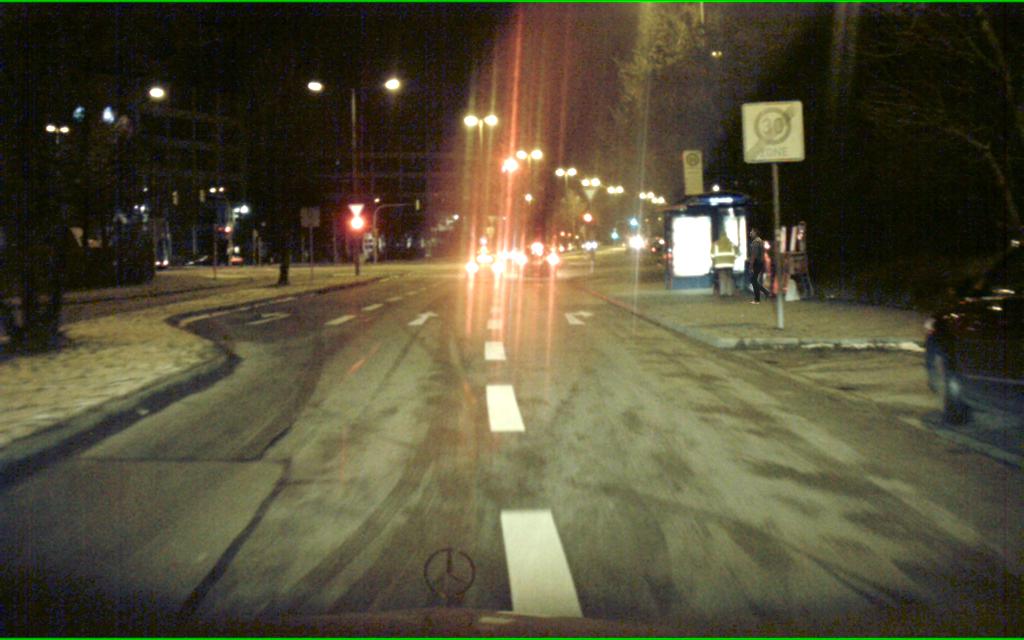}%
\includegraphics[width=0.49\linewidth]{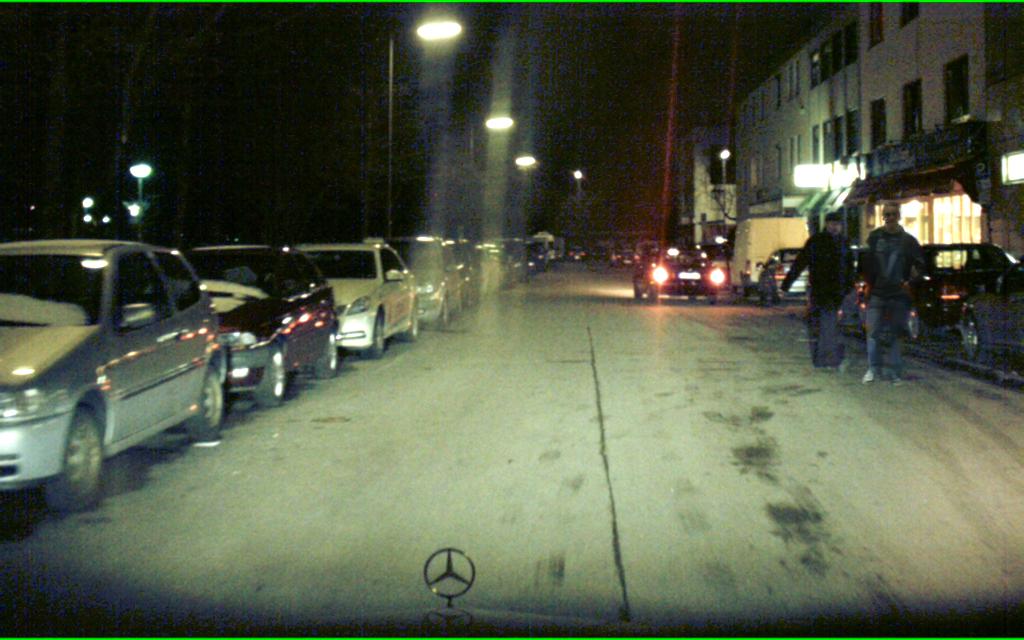}
\end{subfigure}
\begin{subfigure}{\w\linewidth}
\centering
\includegraphics[width=0.49\linewidth]{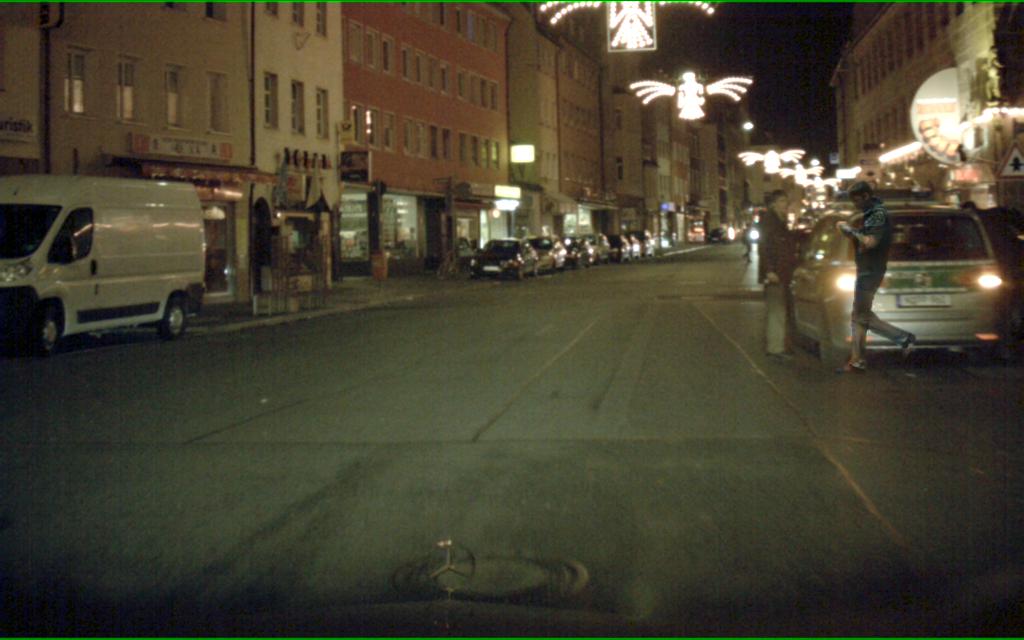}%
\includegraphics[width=0.49\linewidth]{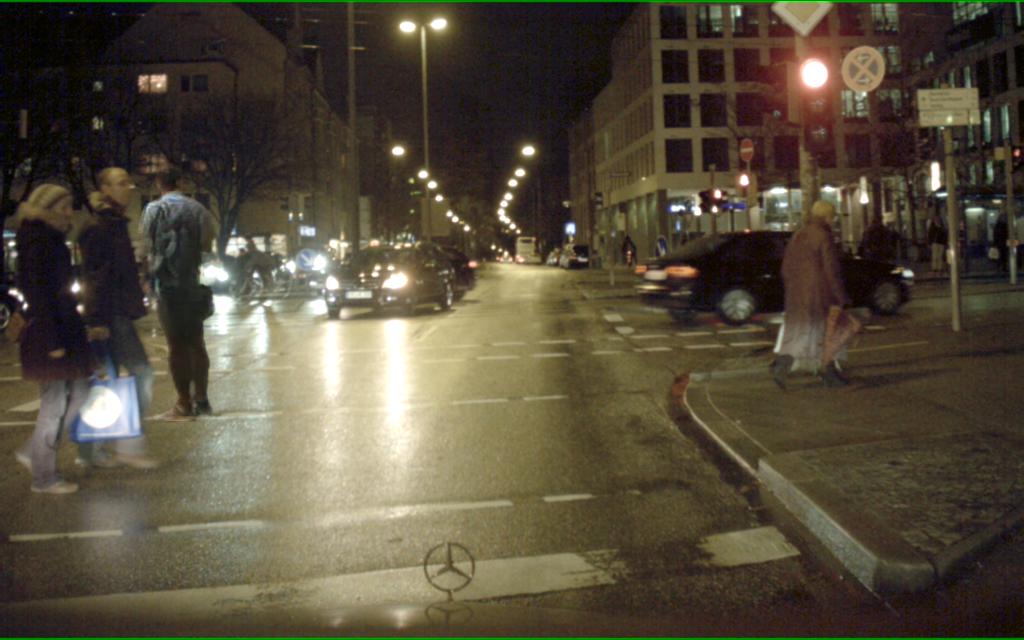}
\end{subfigure}
\caption{{\bf Synthesised samples in NightOwls scenes.} Examples of generated night time images by our DummyNet. One synthetic person was added to each image.}
\label{fig:nightowls_examples_generated}
\end{figure*}

We further generate a broad set of synthetic pedestrians with our DummyNet on person-free night backgrounds and form three synthetic sets of size $5$k, $10$k, and $20$k images. We then perform various experiments with classifiers either trained on synthetic data only or a combination of real and synthetic data. 
We observe that in such a challenging scenario as person detection at night, the synthetic samples bring a considerable improvement. Complete results are show in table~\ref{tab:exp_nightowls_complete}.
Examples of real and synthesized people in NightOwls scenes are shown in Figure~\ref{fig:nightowls_examples_generated}.

\begin{table*}[t]
\setlength{\tabcolsep}{5pt}
\centering
\footnotesize
 \begin{tabular}{c|cc|cc|cc|cc}
 \toprule

gen \textbackslash $\ $ real  & \multicolumn{2}{c|}{$0$} & \multicolumn{2}{c|}{$100$} & \multicolumn{2}{c|}{$1000$} & \multicolumn{2}{c}{$12000$ (full set)} \\
  \midrule
FPR & $1\%$ \em & $10\%$ & $1\%$ & $10\%$ & $1\%$ & $10\%$ & $1\%$ & $10\%$  \\ 
  \midrule
 0 & & & 0.882 & 0.623 & 0.642 & 0.343 & 0.506 & 0.284 \\
 5k & 0.755 & 0.493 & 0.720 & 0.436 & 0.625 & \bf{0.328} & 0.480 & 0.240 \\
 10k & \bf{0.706} & \bf{0.461} & 0.719 & \bf{0.415} & \bf{0.582} & \bf{0.328} & \bf{0.472} & 0.251 \\
 20k & 0.763 & 0.522 & \bf{0.715} & 0.418 & 0.619 & 0.358 & 0.497 & \bf{0.221} \\
\bottomrule 
\end{tabular}
\caption{\small 
	{\bf Night-time person detection on the Nightowls dataset.} We report the mean classifier test set miss rate (lower is better) over five runs. Test results are reported at $1\%$ and $10\%$ FPR. The best results for every combination are shown in bold. Generated samples by our method help to train a better classifier, often by a large margin, over the baseline trained only from real images (the first row corresponding to 0 generated samples). 
}
\label{tab:exp_nightowls_complete}
\end{table*}

\subsection{Improving state-of-the-art person detector}
\label{sec:expC_details}

In table~\ref{tab:citypersons_det_complete}, we give a complete set of results of experiment~C from the main paper. We show a comparison of our method and various baselines in all available setups.

\begin{table*}[th]
\centering 
\footnotesize
\begin{tabular}{p{8cm} | p{1.35cm} | p{1.35cm} | p{1.35cm} | p{1.35cm} | p{1.35cm}}
\toprule 

 setup & reasonable & small & bare & partial & heavy \\
 \midrule
 (a) CSP~\cite{liu2019high} reported & $11.02\%$ & $15.96\%$ & $7.27\%$ & $10.43\%$ & $49.31\%$ \\
 (b) CSP~\cite{liu2019high} reproduced & $11.44\%$ & $15.88\%$ & $8.11\%$ & $10.72\%$ & $49.77\%$ \\
 (c) SURREAL.~\cite{varol17_surreal} aug & $11.38\%$ & $17.39\%$ & $8.06\%$ & $10.56\%$ & $48.24\%$ \\
 (d) CPL~\cite{DwibediMH17} aug. & $11.36\%$ & $16.46\%$ & $7.47\%$ & $10.84\%$ & $48.44\%$ \\
 (e) ADGAN~\cite{Men2020Controllable} aug., orig. & $10.85\%$ & $16.20\%$ & $7.12\%$ & $10.55\%$ & $\mathbf{47.5\%}$ \\
 (f) H3.6M~\cite{h36m_pami} aug., orig. & $11.07\%$ & $16.66\%$ & $7.58\%$ & $10.58\%$ & $50.21\%$ \\
 (g) H3.6M~\cite{h36m_pami} aug., DLv3~\cite{chen2018encoder} & $10.59\%$ & $16.00\%$ & $7.32\%$ & $10.21\%$ & $48.57\%$ \\
 (h) DummyNet aug. (ours) &  $\mathbf{10.25\%}$ &  $\mathbf{15.44\%}$ & $\mathbf{6.95\%}$ & $\mathbf{\:\,9.12\%}$ & $48.60\%$ \\
\bottomrule 
\end{tabular}
\caption{\small \textbf{Improving state-of-the-art person detector~\cite{liu2019high}.} Log-average miss rate of the detector (lower is better) in multiple testing setups provided in~\cite{liu2019high}.}
\label{tab:citypersons_det_complete}
\end{table*}

\subsection{Person detection in night-time scenes}
\label{sec:expD_details}

We choose to train an FPN RCNN with ResNet-50 backbone as the detection network. We initialize the networks with the weights of the network trained for the detection task on the COCO dataset. We train for a total of $50$k iterations with a batch size of $20$, a learning rate of $1e-4$ ($5e-5$ for augmented setups) that is decreased by a factor of $10$ after $25$k and $40$k iterations. We also use a linear warmup for the first $5$k iterations.
Results for all test setups can be found in table~\ref{tab:nightowls_det_complete}.

\begin{table*}[!ht]
\centering 
\footnotesize
\begin{tabular}{p{6.3cm} | p{1.5cm} | p{1.5cm} | p{1.5cm} | p{1.5cm}}
\toprule
 setup & reasonable & small & occluded & all \\
 \midrule
 (a) CityPersons annotations & $41.90\%$ & $50.55\%$ & $65.95\%$ & $50.16\%$ \\
 (b) ADGAN~\cite{Men2020Controllable} aug. & $36.60\%$ & $48.91\%$ & $54.49\%$ & $45.06\%$  \\
 (c) SURREAL~\cite{varol17_surreal} aug. & $32.94\%$ & $44.05\%$ & $49.24\%$ & $41.53\%$ \\
 (d) CPL~\cite{DwibediMH17} aug. & $30.73\%$ & $49.61\%$ & $54.60\%$ & $40.80\%$  \\
 (e) Human3.6M~\cite{h36m_pami} aug. & $27.83\%$ & $43.81\%$ & $45.67\%$ & $36.73\%$ \\
 (f) DummyNet aug. (ours) & $\mathbf{24.95\%}$ & $\mathbf{39.73\%}$ & $\mathbf{44.89\%}$ & $\mathbf{33.69\%}$ \\
\bottomrule 
\end{tabular}
\caption{\small \textbf{NightOwls detection.} Log-average miss rate of the detector (LAMR, lower is better) in multiple testing setups~\cite{Neumann2018NightOwls}.}
\label{tab:nightowls_det_complete}
    \vspace*{-5mm}
\end{table*}

\clearpage

%% file: include/ablations.tex
\section{Ablations}
\label{sec:ablations}
In this section, we show ablations of the individual components of our framework. We evaluate on the NightOwls experiment from the paper (Sec. B). To focus on the usefulness of the generated samples, we compare results in the setup with no real samples. This setup eliminates possible interference caused by real samples. For convenience, Miss Rate (MR) results from the main paper are shown again in the first row of Table~\ref{tab:ablations}.  We conduct ablations on the importance of (i) pose change (Sec.~\ref{sec:abl_pose}), (ii) the importance of the Mask Estimator (Sec.~\ref{sec:abl_mask}), (iii) the importance of conditioning on the pedestrian appearance (Sec.~\ref{sec:abl_app}) \AV{and (iv) the importance of the background influence (Sec.~\ref{sec:ablation_background}) that is studied by an ablation study where all pedestrians are synthesized on a single background then cropped and inserted to different backgrounds in a post-processing step, i.e. the generator does not see different backgrounds as input at test time.} All results are mean values over five runs. 

\begin{table}[!th]
    \setlength{\tabcolsep}{10pt}
    \centering 
    \footnotesize
     \begin{tabular}{l|cc}
     \toprule
     & \multicolumn{2}{c}{MR @ FPR} \\
    Method & $1\%$ \em & $10\%$ \\ 
      \midrule
     (a) ours & \bf{0.706} & \bf{0.461} \\
     (b) fixed pose & 0.891 & 0.766 \\
     (c) non-learned mask & 0.770 & 0.505 \\
 (d) arbitrary appearance & 0.740 & 0.524 \\
     (e) normal distribution & 0.770 & 0.535 \\
     (f) fixed appearance & 0.875 & 0.646 \\
     (g) fixed background & 0.877 & 0.676 \\
    \bottomrule 
    \end{tabular}
    \caption{
    	Comparison of (a) our proposed method and several ablations (b-g). In each row we report an average result over 5 runs. We take the best result obtained using either 5k, 10k or 20k synthetic samples. The numbers represent miss rate (MR) at either $1\%$ or $10\%$ false positive rate (FPR, the lower, the better). Our method achieves the best result overall for both $1\%$ and $10\%$ FPR.
    }
    \label{tab:ablations}
\end{table}

\subsection{Pose Change}
\label{sec:abl_pose}
\paragraph{Fixed Pose.} In this experiment, we focus on the benefits of having the ability to control the pose of the generated pedestrians. We have therefore restricted the system to generate samples in one pose only. Please refer to results in Table~\ref{tab:ablations}, row~(b). This results in a significant decrease in performance, which shows the importance of having the ability to control and vary the pose of the resulting sample. 

\subsection{Importance of the estimated person mask}
\label{sec:abl_mask}
\paragraph{Convex Hull instead of Mask Estimate.} In this experiment, we have studied the benefits of using our Mask Estimator. One alternative to estimating the mask is to use the convex hull of the input keypoints. The results are in Table~\ref{tab:ablations},  row~(c). When compared to the our full approach (row (a)) that uses the Mask Estimator to get more precise silhouettes, we can see an increase of $6.4\%$ in MR at $1\%$ FPR. 
This emphasizes and validates the benefit of using our Mask Estimator.

\subsection{Appearance Conditioning}
\label{sec:abl_app}
In this ablation study, we show the importance of appearance variation when conditioning the generator to improve the quality of the augmented data. Given that we aim to synthesize people that  fit in background scenes of the  NightOwls dataset, the natural choice is to generate darker samples (as it was done in the experiment Sec.\,B in the main paper). We experiment with three alternatives and show that the ability to change the appearance and choose the most fitting appearance to the background is important.

\paragraph{Sampling appearance from images of people.}
The first option is not to restrict the generator conditioning on the dark samples only but obtain the appearance vector from an arbitrary image of a person. The results of this experiment are shown in the Table~\ref{tab:ablations}, row~(d). The best result is reached for the setup with $5$k  synthetic samples in contrast to $10$k in the paper. The performance also decreases by $3.4\%$ which suggests that it is beneficial to have the ability to restrict the appearance to only darker samples as done in the main paper.

\paragraph{Sampling from a Gaussian.}
Since the Person Appearance Encoder was trained as a part of a Variational Autoencoder, its outputs come from the normal distribution $\mathcal{N}\left( \mathbb{0},\mathbb{I} \right)$. Therefore, in the second experiment, we replace the latent appearance vectors obtained from the Person Appearance Encoder by latent appearance vectors randomly drawn from $\mathcal{N}\left( \mathbb{0},\mathbb{I} \right)$. The results are shown in Table~\ref{tab:ablations}, row~(e). The best result is obtained when using $10$k synthetic samples, just as in the main paper; however, the miss rate (MR) at $1\%$ false positive rate (FPR) increases by $6.4\%$ which once again highlights the importance of the ability to directly restrict the appearance to a given set of input images.

\paragraph{Fixed Appearance.}
We experiment with freezing the appearance conditioning to a single appearance sample. The results are shown in the Table~\ref{tab:ablations}, row~(f). This results in a major decrease in the performance compared to all the previous results. This suggests that the variability of the sample appearance is crucial for the success of the augmentation.

\subsection{Background influence}
\label{sec:ablation_background}
\AV{
Finally, we conduct an ablation study where all pedestrians are synthesized on a single background then cropped and inserted to different backgrounds in a post-processing step, i.e. the generator does not see different backgrounds as input at test time.
As it can be seen from the results in the row~(g), this augmentation setup performs significantly worse with log-average-miss-rate (LAMR, lower is better) $87.7$ vs $70.6$ (ours) at $1\%$ FPR and $67.6$ vs $46.1$ (ours) at $10\%$ FPR.
This clearly demonstrates the importance of learning the adaptation of the generated pedestrians to the input backgrounds.
}